\documentclass[runningheads]{llncs}
\usepackage[T1]{fontenc}
\usepackage{graphicx}
\usepackage{times}
\usepackage{soul}
\usepackage{url}
\usepackage[hidelinks]{hyperref}
\usepackage{caption}
\usepackage{graphicx}
\usepackage{amsmath}
\usepackage{amsfonts}
\usepackage{booktabs}
\usepackage{algorithm}
\usepackage{algpseudocode}
\usepackage[switch]{lineno}

\usepackage{subcaption}
\usepackage{siunitx}
\usepackage{multirow}
\begin{document}
\title{\textsc{ImbalancE}: Inference-Time Latent Search Against Degree Imbalance in Link Prediction}
\titlerunning{\textsc{ImbalancE}: Latent Search against Degree Imbalance}
\author{Alberto Bernardi\inst{1}\orcidID{0000-0003-1573-3234} \and Luca Costabello\inst{1}\orcidID{0000-0002-0720-9347} \and Christophe Gueret\inst{1}\orcidID{0000-0002-8914-6107}}
\authorrunning{A. Bernardi et al.}
\institute{Accenture Labs, Dublin, Ireland
%\institute{Princeton University, Princeton NJ 08544, USA \and
\email{\{alberto.bernardi,luca.costabello,christophe.gueret\}@accenture.com}}
\maketitle              % typeset the header of the contribution
\begin{abstract}
    Knowledge Graph Embedding models have been extensively used to learn representations of entities and relations in Knowledge Graphs for predicting missing links. However, the quality of the learned representations varies a lot across different areas of the graph. If previous research has loosely linked the problem to relation types or degree bias, we show that it is more widespread and it correlates with the degree imbalance of the entities in test triples. In particular, the prediction of a target entity that has a degree much smaller than the degree of the anchor entity is extremely problematic.
    %This is critical in use cases like \textit{drug target discovery}, where these triples are predominant, or \textit{recommender systems}, where they represent important corner cases.
    This is critical in \textit{recommender systems} and other use cases, where these triples represent important corner cases.
    To address this issue, we propose an inference-time latent search optimization method capable of significantly improving model predictions on the most imbalanced triples. Built on top of a pre-trained model, it explores the embedding space at evaluation time, blending known and out-of-band information to mitigate the degree imbalance bias. We show the value of our approach on imbalanced triples from common benchmark datasets, where we outperform conventional methods, opening the door to the successful adoption of Knowledge Graph Embedding models on these critical corner cases.
\end{abstract}
\section{Introduction}
Knowledge Graphs (KGs) are flexible and scalable data structures that model factual knowledge by linking concepts through relationships \cite{hogan2021survey}. The possibility of representing and integrating virtually any type of knowledge has made them scale up to include millions or billions of facts. However, such a size is incompatible with manual curation and leaves the door open for incompleteness \cite{dong2014knowledge_vault}. 
In this scenario, the design of a machine-based approach to infer missing links has attracted a lot of attention from the scientific community. This task is known as \textit{link prediction}, where the goal is to predict whether a relation $p$ exists between two entities $s$ and $o$ in the graph. In this context, a \textit{query} takes the form of an incomplete triple, such as $(s, p, ?)$ or $(?, p, o)$, and the model must infer the missing entity. The known entity in the query (e.g., the subject in $(s, p, ?)$) is called the \textit{anchor}, while the missing one is the \textit{target}. The most successful attempts to solve this task have been carried out through Knowledge Graph Embedding (KGE) models, a family of scalable methods that learn low-dimensional representations for entities and relations.

Despite their success, KGEs suffer from limitations, as the quality of the learned representation significantly varies across the graph, limiting the model performance. Previous work has attributed such behavior to relation types \cite{bordes2013transe,wang2014transh,lin2015transr,ji2015transd,he2015kg2e} and degree bias \cite{mohamed2020popularity_bias,shomer2023degree_bias}. In this work, we provide a detailed analysis, identifying the \textit{degree imbalance} of the head and tail entities in test triples as a consistent factor behind this variation. We show how a bigger degree difference between the two correlates with better predictions of the higher degree entity but with much poorer predictions of the lower degree entity. We further tie the problem to a two-fold learning issue that affects low-degree entities, unveiling strong overfitting and failed convergence  behind the unsuccessful predictions.

\begin{figure*}[t]
    \centering
    \makebox[\textwidth][c]{%
    \includegraphics[width=1.2\linewidth]{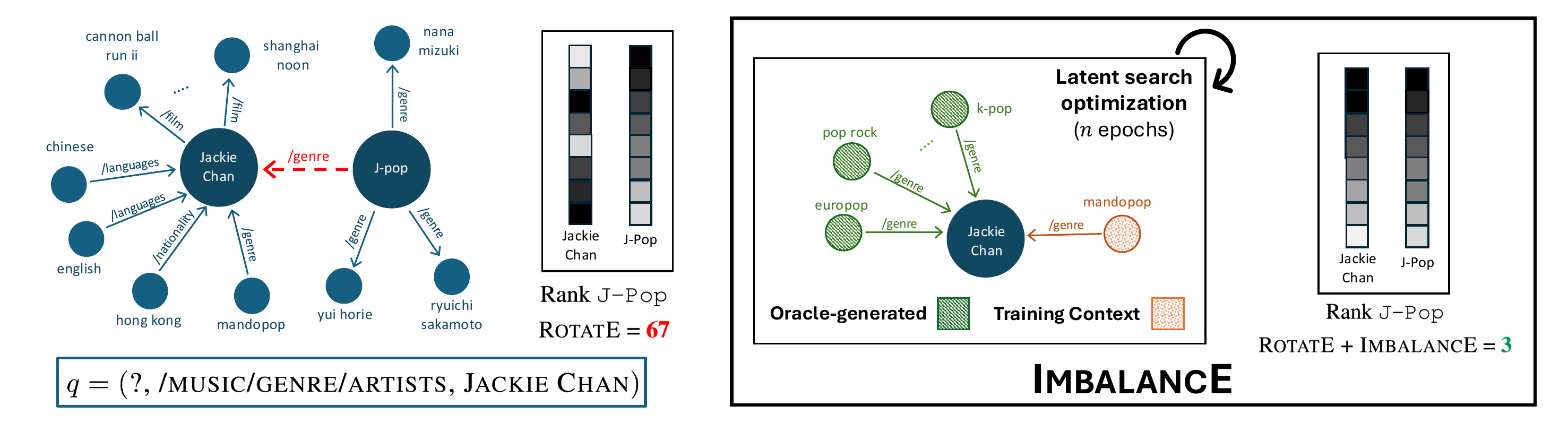}
    }
    \caption{We consider a query $q$ with high degree imbalance. In this example taken from FB15k-237, \textsc{Jackie Chan} has degree 49 and \textsc{J-Pop} has degree 8 (limited in the figure for clarity of visualization). \textsc{ImbalancE} takes the pre-trained embeddings of the KGE model for the anchor entity and performs an inference-time latent search optimization on a subset of the original neighborhood enhanced with the triples generated by the oracle. It then outputs a representation for the anchor node that better aligns with the selected query, improving the ranking of the target answer from 67 down to 3.}
    \label{fig:hero_image}
\end{figure*}

In \textit{recommender systems}, link prediction has been deployed frequently \cite{zhang2016kg_for_recommender}. On a music streaming platform, we want to complement artists metadata as much as possible to make recommendations more accurate. 
Consider \textsc{(J-pop, /music/genre/artists, Jackie Chan)}, a test triple from FB15k-237 \cite{toutanova2015fb15k237}. The degree of \textsc{J-pop} is low, while \textsc{Jackie Chan} has a much higher number of connections. We want to answer the query $q = \textsc{(?, /music/genre/artists, Jackie Chan)}$, to tag \textsc{Jackie Chan} with an additional and related genre to broaden his audience. When using embeddings from a pre-trained KGE model, the predicted rank for \textsc{J-pop} is poor because of the large degree difference between \textsc{Jackie Chan} and \textsc{J-pop} (see Figure~\ref{fig:hero_image}).

The problem affects a broad range of triples and various Graph Machine Learning models: from shallow architectures like TransE, DistMult, ComplEx and RotatE, to GNN-like ones, including the state-of-the-art NBFNet.

% Latent search method presentation.
To address the issue, we propose \textsc{ImbalancE}, an inference-time latent search method.  Given the pre-trained embeddings of a KGE model and an imbalanced test triple $t = (s, p, o)$, \textsc{ImbalancE} improves the prediction of the low-degree entity by fine-tuning the embedding of the anchor node and of few selected entities optimizing a dual-term objective function. These two terms extend the generalization of the embedding of the high-degree node and improve the quality of the embedding of the low-degree nodes. The former goal is achieved by an exploration of the embedding space guided by a selection of training facts: this second pass ring-fences the representation of the high-degree node in a region of the space suitable for the prediction. The latter goal, on the other hand, is achieved through an oracle that extends the little knowledge available in the KG for long-tail entities.

We validate our approach on the most imbalanced triples of common benchmark datasets. We show how \textsc{ImbalancE} leads to significant improvements, enabling the prediction of low degree entities in critical applications where these corner cases are predominant. 

% Contributions
In summary, our work makes the following contributions:
\begin{itemize}
    \itemsep-0.05em 
    \item[1.] We detect and analyze the degree imbalance affecting a variety of KGE models, and we further tie the problem to a two-fold learning issue that compromises the embedding quality of low-degree entities.
    \item[2.] We propose \textsc{ImbalancE}, an inference-time latent search method that mitigates the above problem. We assess its impact on heavily imbalanced triples in popular \textit{link prediction} benchmarks across multiple KGE models and show that it works as an effective plug-in to enhance KGE predictions. Moreover, we show how it can provide a lightweight and efficient alternative to much more involved and computationally expensive models.
\end{itemize}
% Structure of the paper
After the related work (Section~\ref{sec:related}), the paper introduces the degree imbalance problem and its implications in Section~\ref{sec:degree_imbalance}. Section~\ref{sec:latent_search} describes \textsc{ImbalancE}, from the intuition to the technical details, while Section~\ref{sec:experiments} presents the experimental results. We draw conclusions, limitations, and future directions in Sections~\ref{sec:limitations} and \ref{sec:conclusions}.

\section{Related Work}
\label{sec:related}

\subsubsection{Knowledge Graph Embedding Models} KGE models learn continuous representations of entities and relations in the KG from the graph topology and its soft regularities. A long list of methods \cite{cao2024survey_kge} has followed seminal work in the space \cite{nickel2012factorizing}. In this work, we limit our analysis to four traditional KGE models: TransE \cite{bordes2013transe}, DistMult \cite{yang2015distmult}, ComplEx-N3 \cite{trouillon2016complex,lacroix2018complex_n3} and RotatE \cite{sun2019rotate}. Although others have claimed better performance \cite{balazevic2019tucker}, the gain is often marginal, and the small differences in modeling the representation of triples are not relevant to the core contribution of our work. A notable exception is represented by NBFNet \cite{zhu2021neural}, a GNN-like architecture that has achieved substantial improvements. Nonetheless, it also suffers from degree imbalance (see Appendix~\ref{apx:degree_imbalance_more_model}), but it cannot benefit from \textsc{ImbalancE}. In fact, the entity embeddings are heavily entangled within the message-passing architecture. This prevents the independent update of the anchor and other target entities, while freezing the relation embedding, making the approach not scalable and prone to overfitting.

\subsubsection{Topology-related Issues on KGs} The topology of KGs introduces significant challenges in training KGE models \cite{sardina2024surveyknowledgegraphstructure}. \cite{mohamed2020popularity_bias} and \cite{rossi2021relation_bias} have exposed the bias toward high-degree nodes and how aggregate metrics can offer a distorted view of the actual performance of models. However, they loosely defined the issue and did not propose any countermeasure. \cite{shomer2023degree_bias} focused on the frequency of entity-relation pairs, proposing the synthetic generation of additional embeddings to compensate for long-tail entities distribution. Other works \cite{bordes2013transe,ji2015transd,lin2015transr,he2015kg2e} have identified the heterogeneity of relation types (in particular, 1-to-many and many-to-1) as one of the main obstacles. They have altered the representation of relations toward a more expressive one.
Our work surpasses all these previous efforts as it defines the problem in a more systematic way, tracing it back to a two-fold learning issue of the training, extending the scope to a broader set of triples, and proposing an effective solution.

\subsubsection{Latent Search Optimization}
The application of an inference-time latent search optimization has been used to explore latent program spaces \cite{bonnet2024searchinglatentprogramspaces}. To the best of our knowledge, our work is the first to leverage such a latent search optimization for KGE models, which requires non-trivial adaptations. Refer to the work above for an overview of other application domains that exceed the scope of our work.

\section{What Aggregate Metrics Do not Show}
\label{sec:degree_imbalance}

KGE models have proven to be successful in reconstructing missing links in KGs. However, aggregated metrics often hide that the quality of these predictions varies enormously in different groups of triples. Previous work has loosely associated the problem with the epistemic uncertainty of low-degree entities and has shown the challenge of predicting head (and tail) of 1-to-N (and N-to-1) relations. We extend the scope of the issue, proving that a major driver of prediction difficulty lies in the degree difference (or \textit{degree imbalance}) between the subject and the object: this makes the reconstruction of the higher degree entity much simpler, while leaving the lower degree entity poorly reconstructed. 

In Figure~\ref{fig:mrr_degree_diff}, we plot the Mean Reciprocal Rank (MRR) of ComplEx and RotatE (more models in Appendix~\ref{apx:degree_imbalance_more_model}), isolating the performance on subject and object corruptions, and binning the test triples of FB15k-237 \cite{toutanova2015fb15k237} and Yago3-10 \cite{mahdisoltani2015yago3} based on the normalized degree difference, that we define as:
\[
\hat{\Delta}(t = (s, p, o)) = \frac{\delta(s) - \delta(o)}{\min \left\lbrace \delta(s), \delta(o) \right\rbrace},
\]
where the degree $\delta$ counts incoming and outgoing edges of a node.
We can immediately observe the stark difference in the quality of the predictions for highly imbalanced triples and how this gap decreases as the normalized degree difference $\hat{\Delta}$ approaches zero. Although the aggregate metrics reported in Table~\ref{tab:kge_agg_metrics} portray models with robust predictive power, Figure~\ref{fig:mrr_degree_diff} shows that metrics are inflated by easier predictions, while these models could hardly be trusted in use cases where the prediction of low-degree entities is predominant.

\begin{table}[tb]

\centering
\setlength{\tabcolsep}{4pt}
\caption{Aggregate performance of KGE models on FB15k-237, WN18RR, and YAGO3-10.}
\begin{tabular}{c c c c c c c}
 & \multicolumn{2}{c}{\textbf{FB15k-237}} & \multicolumn{2}{c}{\textbf{WN18RR}} & \multicolumn{2}{c}{\textbf{YAGO3-10}} \\
\cmidrule(lr){2-3} \cmidrule(lr){4-5} \cmidrule(lr){6-7}
 & \textbf{MRR} & \textbf{H@10} & \textbf{MRR} & \textbf{H@10} & \textbf{MRR} & \textbf{H@10} \\
\midrule
ComplEx   & 0.31 & 0.49 & 0.51 & 0.58 & 0.36 & 0.56 \\
RotatE    & 0.31 & 0.51 & 0.52 & 0.61 & 0.37 & 0.57 \\
\bottomrule
\end{tabular}

\label{tab:kge_agg_metrics}
\end{table}

\begin{figure*}[t]
    \centering
    
    % --- FB15k-237 ---
    \begin{subfigure}{\linewidth}
        \centering
        \includegraphics[width=0.7\linewidth]{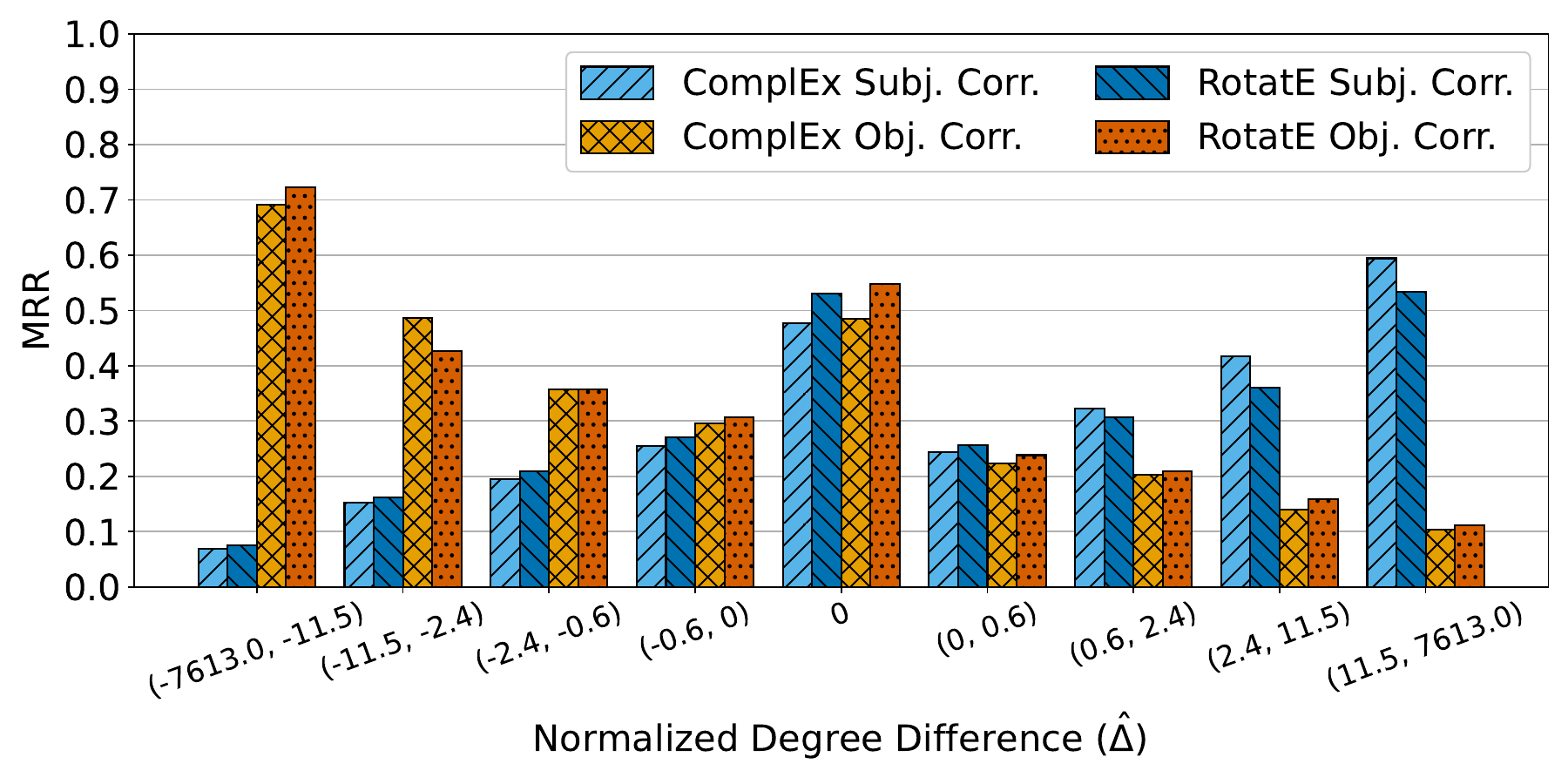}
        \caption{FB15k-237}
        \label{fig:mrr_fb15k237}
    \end{subfigure}
    
    \vspace{0.5cm}
    
    % --- YAGO3-10 ---
    \begin{subfigure}{\linewidth}
        \centering
        \includegraphics[width=0.7\linewidth]{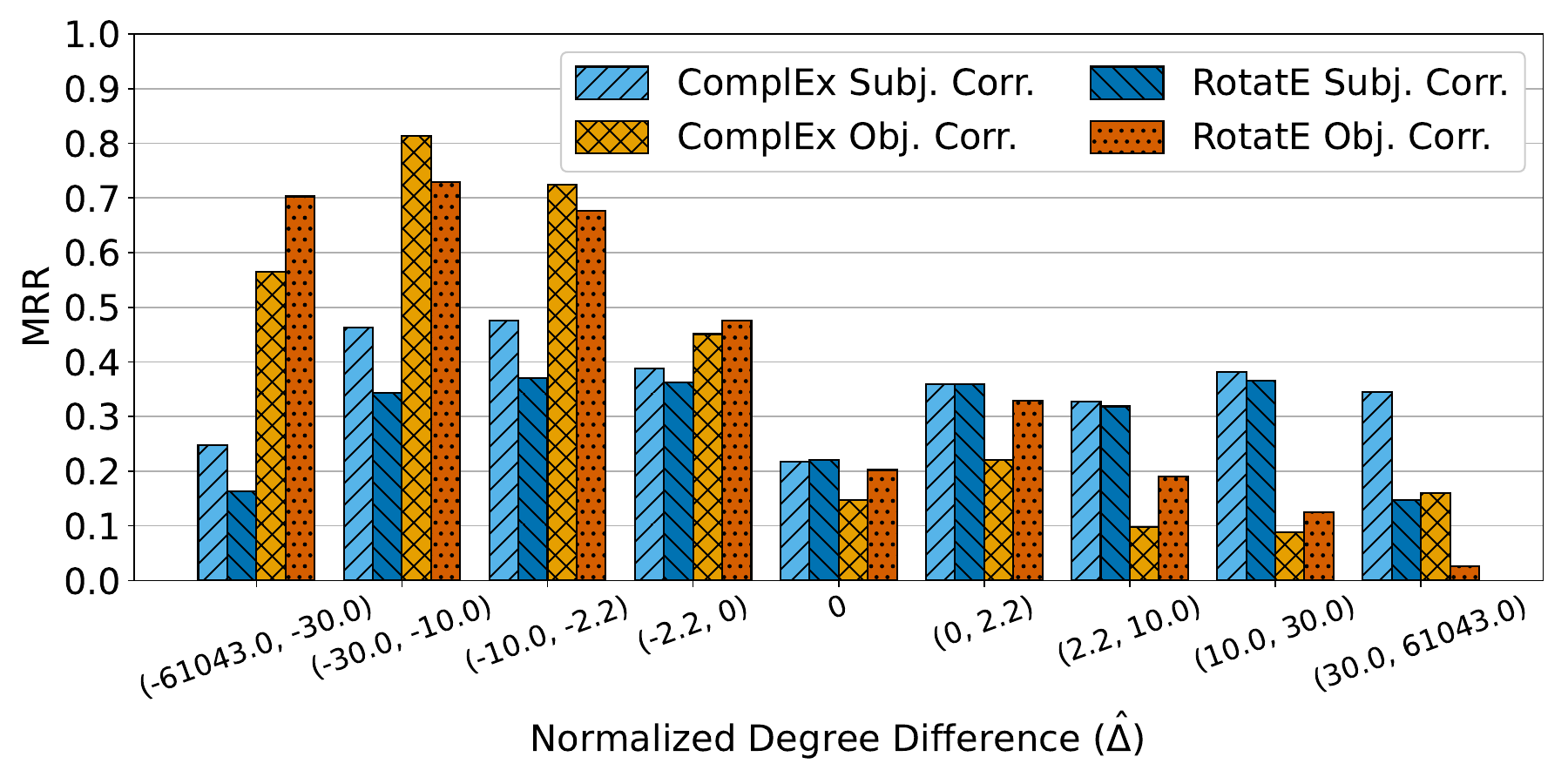}
        \caption{YAGO3-10}
        \label{fig:mrr_yago310}
    \end{subfigure}

    \caption{Performance of conventional KGE models on test triples binned by normalized degree difference $\hat{\Delta}$, using quartiles of $\lvert \hat{\Delta} \rvert$. On the left hand side of each plot we have triples with high-degree object and low-degree subject, on the right hand side, high-degree subjects and low-degree objects. As the degree imbalance increases, the performance gap between high-degree and low-degree prediction increases.}
    \label{fig:mrr_degree_diff}
\end{figure*}

To explain the degree imbalance issue, we take a step back and look into the training process. We carry out an analysis on FB15k-237 and report the results obtained for RotatE. % TODO:, but similar results hold for different models and datasets. %Refer to Appendix~\ref{apx:convergence_issue} for all the technical details and similar results for different models and datasets.

In conventional KGE models, entities and relations are assigned a low-dimensional, continuous representation to capture relational patterns in the graph structure. During every training epoch, these representations are optimized to maximize the score assigned to the ground truth triples in the KG (\textit{positives}), and minimize the score for false statements (\textit{negatives}). In particular, the embedding of an entity is moved around the latent space every time the model processes a positive or a negative involving that entity.

\begin{figure*}[t]
\centering
% Top plot
\begin{subfigure}[t]{0.8\textwidth}
\small
    \centering
    \makebox[\textwidth][c]{%
    \includegraphics[width=1.2\linewidth]{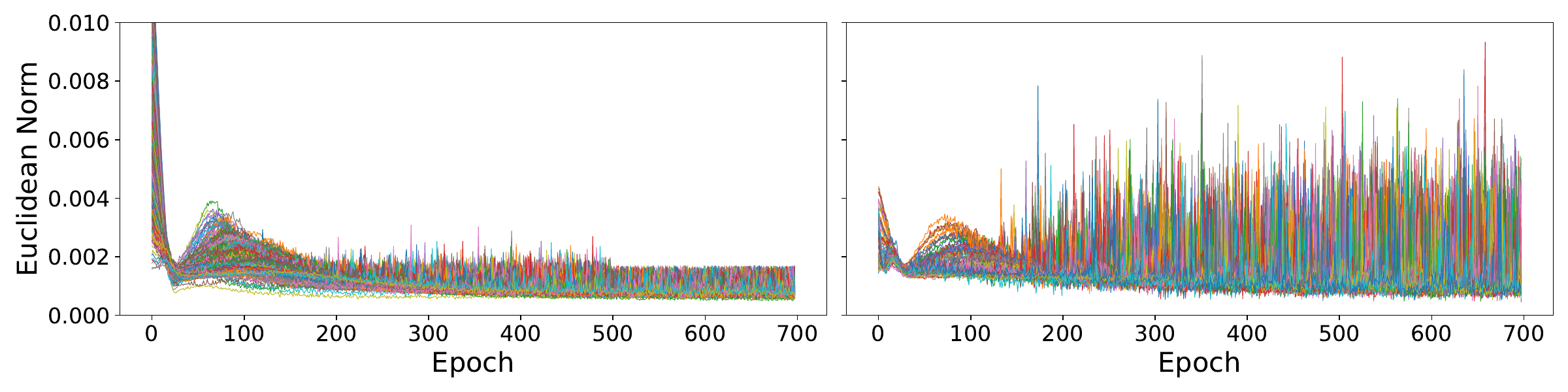}
    }
    \caption{On the left, entities converge and the update reduces to approach zero. On the right, embeddings of the entities keep getting updated, proving there is an issue with the learning process.}
    \label{fig:grad_updates}
\end{subfigure}

\vspace{1em}

% Bottom row with two side-by-side subfigures
\makebox[\textwidth][c]{%
\begin{subfigure}[t]{0.52\textwidth}
    \centering
    \includegraphics[width=\linewidth]{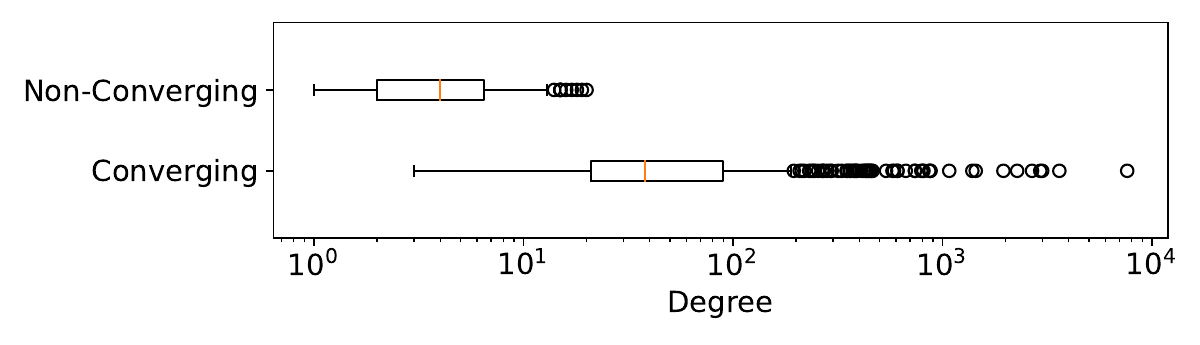}
    \caption{Degree distribution}
    \label{fig:degree_groups}
\end{subfigure}
\begin{subfigure}[t]{0.52\textwidth}
    \centering
    \includegraphics[width=
    \linewidth]{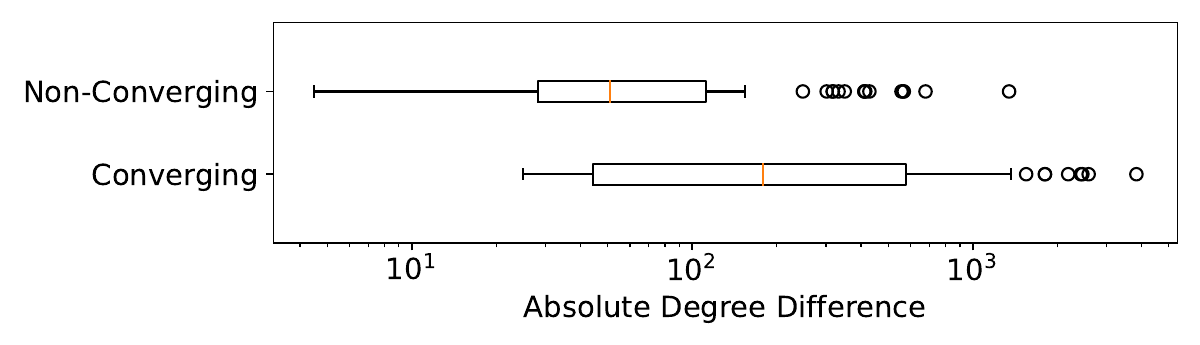}
    \caption{Absolute degree difference ($\lvert \Delta \rvert $)}
    \label{fig:diff_deg_groups}
\end{subfigure}
}
\caption{(Top) Euclidean Norm of the difference between embeddings of converging and non-converging entities while training RotatE on FB15k-237. (Bottom) Comparison of degree statistics for converging vs.~non-converging entities.}
\label{fig:all_plots}
\end{figure*}

By plotting the (euclidean) distance between the embeddings of the same entity across successive epochs, we can get an idea of the magnitude of the updates and, as it decreases, of the rate of convergence of the learning process (Figure~\ref{fig:grad_updates}). We observed that for one group of entities the magnitude of the updates stabilizes to very low values, indicating convergence. Surprisingly, a second group keeps receiving significant updates, with randomly repeating peaks indicating that the node embedding keeps moving around the space. Such a behavior suggests a failed convergence, rather than an insufficient amount of training, as entities keep getting consistent updates over and over again.

Digging deeper into the characteristics of the two groups, we found a significant difference in the degree distributions, with the convergent group having a much higher degree (Figure~\ref{fig:degree_groups}).

To reduce confounding factors, we restricted the subset of converging to the entities with lower degree. In this way, we got similar degree distribution between the two groups. Comparing the degree imbalance of the \textit{training} triples involving converging and non-converging entities,  we noticed how a higher imbalance favors convergence (Figure~\ref{fig:diff_deg_groups}). Paired with the observed difficulty of predicting degree imbalanced triples at test-time observed in Figure~\ref{fig:mrr_degree_diff}, this tells us that apparent convergence of low-degree nodes can be better interpreted as overfitting. 
This is the case for \texttt{J-Pop} in our example above: it converges (i.e., the embedding position stabilizes) as only one relation type characterizes its neighborhood, but the embedding is actually overfitting, as the poor rank assigned to the entity in Figure~\ref{fig:hero_image} shows.
On the contrary, low-degree nodes connected to lower degree entities keep moving around the space, failing to converge, and posing an equally hard challenge for the model at test-time in light of their instability. 

Therefore, the challenge underlying the most imbalanced triples in the test set is two-fold: a strong overfitting on one side and an extremely poor learning on the other. We set to address this issue by proposing \textsc{ImbalancE}, an inference-time latent search method that tries to improve model predictions on the most imbalanced triples involving low-degree entities.

\section{Inference Time Latent Search for Degree Imbalance}
\label{sec:latent_search}

\subsection{Preliminary and Notation}
A Knowledge Graph $\mathcal{G}=\{ (s,p,o) \} \subseteq \mathcal{E} \times \mathcal{R} \times  \mathcal{E}$ is a set of triples $t=(s,p,o)$, each including a subject (\textit{head}) $s \in \mathcal{E}$, a predicate $p \in \mathcal{R}$, and an object (\textit{tail}) $o \in \mathcal{E}$, where $\mathcal{E}$ and $\mathcal{R}$ are the sets of all entities and relation types, respectively. We refer to the task of predicting unseen triples in a KG as \textit{Link Prediction}. It is formalized in the literature as a learning-to-rank problem, where the objective is learning a scoring function $f: \mathcal{E} \times \mathcal{R} \times \mathcal{E} \rightarrow \mathbb{R}$ that, given an input triple $t=(s,p,o)$, assigns a score $f(t) = f((s,p,o)) \in \mathbb{R}$ proportional to the likelihood that the fact $t$ is true.

\subsection{Intuition}

At the end of training of a conventional KGE model, every entity and relation has a low-dimensional, continuous representation which is the result of updates based on the triples in the KG. As such, it is the representation that best fits \textit{all} training triples. Given a query $q$, however, a one-size-fits-all representation of the anchor might not be precise enough to find the correct answer, especially if the anchor has high degree, which could thus introduce a lot of noise into the representation. Consider again the example above and the node \textsc{Jackie Chan}. All the relations in its neighborhood are different from \textsc{/music/genre/artists} except for one. As a result, the embedding will have little pertinence to the query. This is the reason why we refine this embedding using the training triples that share the anchor and the predicate of $q$, i.e., \textsc{(Mandopop, /music/genre/artists, Jackie Chan)}. In this way, the new representation is the one that best suits $q$ and not \textit{all} triples in the KG. We call this set of triples the \textit{training context} of $q$ and denote it $\mathcal{C}_q = \left\lbrace (s, p, o_i) \vert o_i \in \mathcal{E} \wedge (s, p, o_i) \in \mathcal{G} \right\rbrace$ if $q = (s, p, ?)$. Analogously, if $q = (?, p, o)$, $\mathcal{C}_{q} = \left\lbrace (s_i, p, o) \vert s_i \in \mathcal{E} \wedge (s_i, p, o) \in \mathcal{G} \right\rbrace$.

The second issue of the training optimization process is related to low-degree entities. As we saw in the previous section, their embeddings are either heavily dependent on the few entities they are connected to and thus are very prone to overfitting, or they suffer from an incomplete learning. This makes their reconstruction from the anchor node $s$ very unlikely, as their representation shares too little with $s$. The node \textsc{J-Pop}, for example, is only connected to musicians, while \textsc{Jackie Chan} is best known for his acting. Similarly, other possible genres that are plausible answers for the query could be far off from \textsc{Jackie Chan} for the same reason. Therefore, it is essential to bias their embeddings in favor of the query. To do so, we leverage an external oracle that has the advantage of being degree agnostic, to limit the risk of neglecting low-degree entities. This oracle suggests triples similar to those in the training context of the query. In this way, we will bring the embeddings of \textsc{Jackie Chan} and musical genres similar to \textsc{Mandopop} closer together, making a correct answer to the query more likely.

\subsection{ImbalancE}
Inference-time latent search has been successfully applied to improve generalization and model predictions in a completely different domain by \cite{bonnet2024searchinglatentprogramspaces}. As the degree imbalance could benefit from these properties, we design an inference-time latent search that can be selectively applied to link prediction queries (see Algorithm~\ref{alg:latent_search}). Built on top of a pre-trained KGE model with scoring function $f$, it takes a query $q = (s, p, ?)$ (analogous for $(?, p, o)$) and refines the representation of relevant entities optimizing for $T$ epochs a two-term objective function:
\[
\mathcal{L}(q = (s, p, ?)) =
\sum_{t^+ \in \mathcal{C}_q} f(t^+)
+
\sum_{t \in \Omega_q} f(t). %= \sum_{(s, p, o_i) \in \mathcal{C}_q} f(s, p, o_i) + \sum_{(s, p, o_j) \in \Omega_q} f(s, p, o_j)
\]

\begin{algorithm}[t]
\small 
\caption{Inference-Time Latent Search for Degree Imbalance}
\label{alg:latent_search}
\begin{algorithmic}[1]
\Require Pretrained KGE model with scoring function $f$ and entity embedding matrix $\mathbf{E}$, a query $q = (s, p, ?)$, training context $\mathcal{C}_q$ and oracle triples $\Omega_q$, number of latent search iterations $T$
\Ensure Ranked list of entities
\State $\mathbf{E}_0 \gets \mathbf{E}$
\State Set the embeddings of $s$ to $\mathbf{s}^0 \gets \mathbf{E}_0[s]$
\State Set the embeddings of $o_j \in \left\lbrace o | (s, p, o) \in \Omega_q \right\rbrace$ to $\mathbf{o}_j^0 \gets \mathbf{E}_0[o_j]$, $\forall j=1,...,\lvert \Omega_q \rvert$ 
\For{iteration $=0$, \dots, $T$}
    \State Compute $\mathcal{L}(q)$ and gradients w.r.t. $\mathbf{s}^i$ and $\mathbf{o}_j^i$, $j=1,...,\lvert \Omega_q \rvert$ 
    \State Update $\mathbf{s}^{i+1} \gets \textsc{GradientUpdate}(\mathbf{s}^i)$
    \State Update $\mathbf{o}_j^{i+1} \gets \textsc{GradientUpdate}(\mathbf{o}_j^i)$, $\forall j=1,...,\lvert \Omega_q \rvert$
    \State Update $\mathbf{E}_{i+1}[s] \gets \mathbf{s}^{i+1}$ and $\mathbf{E}_{i+1}[o_j] \gets \mathbf{o}_j^{i+1}$, $\forall j=1,...,\lvert \Omega_q \rvert$
\EndFor
\State Evaluate $q$ with the updated embeddings and extract ranked list of candidates
\State \Return Ranked candidates based on final scores
\end{algorithmic}
\end{algorithm}

\subsubsection{Training Context Term} The first term of the loss function sums the scores assigned to the training context of the query. These are ground truth triples already processed during training, that the model should have already learned. However, as we observed above, the final embedding assigned to the anchor node is a one-size-fits-all representation that could under-perform on a specific query. Therefore, during our latent search, we present these relevant triples again, to better tailor the anchor representation for the final prediction. Importantly, in this term we optimize \underline{only} the embedding of the anchor node of the query, while keeping the others frozen. If we were to optimize the embeddings of head, tail and predicate, we would overfit learned triples even more, thus worsening the already poor generalization of the model at test-time. Limiting the optimization to the anchor node, on the contrary, we refine its representation to best suit the training context, improving its generalization and ring-fencing it in an area of the embedding space suitable for the query.

\subsubsection{Oracle-Enhanced Term} If the first term improves the representation of the anchor, we might still get bad predictions if the embeddings of the query answers are inaccurate. This is particularly likely for low degree entities that are dependent on very few facts that could well be irrelevant for the target query $q$, or, as shown before, prone to overfitting or susceptible to a failed learning. Therefore, the sole refinement of the anchor using the training context is not enough. We also need to adjust the target embeddings, biasing them toward the query. Doing so is not trivial since, at test-time, we have no sense of what the correct answers to the query are. Moreover, being our focus on low-degree targets, little information about them is available within the KG. For this reason, we enhance our latent search with a set of additional triples $\Omega_q = \{(s, p, o_i) \vert o_i \in \mathcal{E}\}$ generated by an oracle that considers them likely answers to $q$. 
This time around, we optimize both the anchor and the target entity embeddings, so that, following the oracle leads, also the landscape of the answer representations changes in favor of the query. \newline
As we said, the KG provides limited information about the target entity. Therefore, our oracle can be any out-of-band source of information well aligned with the task at hand. For our experiments on benchmark datasets we have used a Large Language Model (LLM) defining similarity based on the encoded textual descriptions of the entities (Appendix~\ref{apx:oracle}).
In the recommender system example, the oracle could also be formalized as user interaction patterns, where the similarity between two genres is measured as the ratio of shared listeners. In other real use case scenarios, the oracle could be represented by any form of expertise that extends the structural knowledge encoded in the KG.

A final observation is needed here: the oracle triples may include false positives that could surface up in the evaluation ranking. However, this risk is mitigated by the regularization effect it has on the representation of the anchor and by biasing the representation of possible targets in favor of the query.

\paragraph{} Crucially, our approach does not require any negatives. In fact, freezing the relation and the training context target embeddings prevents the embedding collapse, acting as a regularizer that removes the need for a contrastive term in the loss. This circumvents the problem of synthetic negative generation that has attracted a lot of attention for its multiple criticalities \cite{kamigaito2022survey,madushanka2024survey} and limits the computational overhead of \textsc{ImbalancE}, that is extremely scalable (see Section~\ref{sec:results}). Moreover, it is remarkably flexible, as it can be selectively applied to single queries, allowing to refine single predictions.

\section{Experiments}
\label{sec:experiments}

\subsection{Experimental Settings}

\subsubsection{Datasets and Test Splits.}
We evaluate \textsc{ImbalancE} on three encyclopedic benchmark KGs, widely used in the literature: FB15k-237 \cite{toutanova2015fb15k237}, WN18RR \cite{dettmetters2018conve} and YAGO3-10 \cite{mahdisoltani2015yago3}. To prove its efficacy on highly imbalanced triples involving low-degree entities, we identify \textit{low-degree} nodes as those with degree below the first quartile of the degree distribution, and \textit{high-degree} nodes as those with degree above the third quartile. We then assign to the High-Low (Low-High) split all test triples with the subject having high (low) degree and the object low (high) degree. The statistics of the resulting splits are reported in Table~\ref{tab:stats_splits}, while additional statistics about the datasets are reported in Appendix~\ref{apx:dataset_stats}.

\begin{table}[t]
\centering
    \caption{Statistics of the datasets.}
    \begin{tabular}{lccc}
        \toprule
         & \textbf{FB15k-237} & \textbf{WN18RR} & \textbf{Yago3-10} \\
         \midrule
         \textbf{\#H Nodes (Min degree)} & \num{3536} (\num{41}) & \num{2296} (6) & \num{28913} (5)\\
         \textbf{\#L Nodes (Max degree)} & \num{4015} (\num{11}) & \num{32697} (10) &  \num{31487} (16) \\
         \textbf{\#Valid H-L / L-H} & \num{338} / \num{775} & \num{295} (689) &  \num{25} / \num{298} \\
         \textbf{\#Test H-L / L-H} & \num{396} / \num{942} & \num{277}  / \num{753} & \num{53} / \num{270} \\
         \bottomrule
    \end{tabular}
    \label{tab:stats_splits}
\end{table}

\subsubsection{Evaluation Protocol and Metrics.}
We evaluate \textsc{ImbalancE} by corrupting the low-degree entity of the test triples with all entities in the KG and we consider the \textit{filtered} setting \cite{bordes2013transe}, i.e., we filter from the corruptions all facts in the training, validation or test sets. We then rank test triples against all corruptions. The metrics used are the usual ones for link prediction: Mean Reciprocal Rank (MRR) and Hits at N (Hits@N).

\subsubsection{Hyperparameter-Search} Given a pre-trained KGE model, \textsc{ImbalancE} has only three hyper-parameters: the learning rate $\lambda$ of an Adam \cite{bengio2015adam} optimizer, the number of latent search iterations $T$, and the number of oracle-generated triples included in $\Omega_q$. We carried out a grid search exploring $\lambda \in \{1e-2, 1e-3, 1e-4\}$, $\lvert \Omega_q \rvert \in \{3, 5, 7, 10, 20, 30, 40, 50\}$, and we set $T = 30$, enabling early stopping, and selecting optimal values based on the validation MRR.

\subsection{Results}
\label{sec:results}
\begin{figure}
    \centering
    \includegraphics[width=0.65\linewidth]{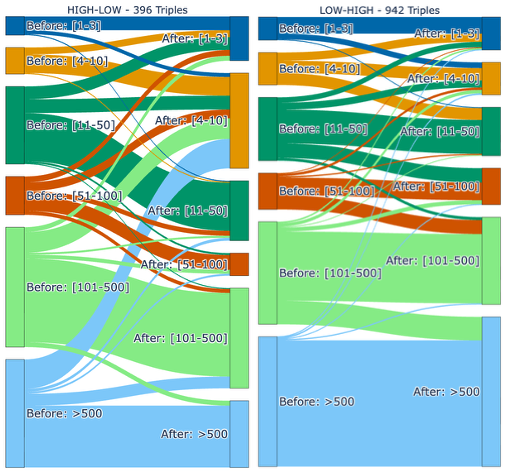}
    \caption{Sankey plots showing the distribution of test triples ranks assigned by RotatE (before) and how the distribution changed applying \textsc{ImbalancE} (after). \textsc{ImbalancE} greatly improves ranks even from far below the ranking ($>$500) bringing them in the top 10.}
    \label{fig:sankey_plot_rotate}
\end{figure}

The results of our experiments are reported in Table~\ref{tab:results}. The impact of the latent search on FB15k-237 triples is striking, with metrics that improved at least by a factor of 2 on the Low-High split and over a factor of 6 on the High-Low one. This shows how \textsc{ImbalancE} compensates the flaws of the pre-trained models. Also on WN18RR there is a noticeable improvement across all metrics, but it is smaller than on FB15k-237. The reason for it is that in WN18RR the degree distribution is concentrated around values much smaller than those of FB15k-237 (Table~\ref{tab:stats_splits}). This reduced polarization results in a less pressing degree imbalance issue, making the improvement of \textsc{ImbalancE} less striking. Finally, on Yago3-10, we only see an improvement on the Hits@10. This behavior can be justified by the fact that the oracle generated the triples for this dataset using only the labels of the entities of the KG, while it leveraged labels and additional descriptions for FB15k-237 and WN18RR. This impacted negatively the quality of triples extracted by the oracle, preventing the correct answers from reaching the top positions of the ranking. Importantly, the impact of \textsc{ImbalancE} goes beyond what metrics show: in fact, the method significantly improves the ranks of triples deep down in the ranking (Figure~\ref{fig:sankey_plot_rotate}).

\begin{table*}[t]
    \centering
    \scriptsize
    \setlength{\tabcolsep}{6pt}
    \caption{Results on the three benchmark datasets on the High-Low and Low-High triples splits. The best value for each metric on each dataset is reported in \textbf{bold} except in case of a tie, when they are \underline{underlined}.}
      \begin{tabular}{llcccccccccc}
        \toprule
         \multirow{2}{*}{\textbf{Dataset}} & \multirow{2}{*}{\textbf{Model}} & \multicolumn{4}{c}{\textbf{High-Low}} & \multicolumn{4}{c}{\textbf{Low-High}}\\
         \cmidrule(lr){3-6} \cmidrule(lr){7-10}
         & & \textbf{MRR} & \textbf{H@1} & \textbf{H@3} & \textbf{H@10} & \textbf{MRR} & \textbf{H@1} & \textbf{H@3} & \textbf{H@10} \\
        \midrule
         \multirow{4}{*}{\textbf{FB15k-237}} & \textbf{ComplEx-N3} & 0.03 & 0.01 & 0.01 & 0.06 & 0.03 & 0.01 & 0.02 & 0.08 \\
         & \textbf{+\textsc{ImbalancE}} & \underline{0.13} & 0.05 & \textbf{0.14} & 0.28 & \underline{0.08} & \underline{0.04} & 0.08 & \underline{0.17} \\
         \cmidrule(lr){3-6} \cmidrule(lr){7-10}
         & \textbf{RotatE} & 0.02 & 0.01 & 0.01 & 0.05 & 0.04 & 0.01 & 0.04 & 0.09 \\
         & \textbf{+\textsc{ImbalancE}} & \underline{0.13} & \textbf{0.06} & 0.11 & \textbf{0.36} & \underline{0.08} & \underline{0.04} & \textbf{0.09} & \underline{0.17} \\
         \midrule
         \multirow{4}{*}{\textbf{Yago3-10}} & \textbf{ComplEx-N3} & 0.06 & 0.00 & 0.03 & 0.23 & 0.07 & \underline{0.02} & 0.07 & 0.17 \\
         & \textbf{+\textsc{ImbalancE}} & 0.06 & 0.00 & 0.03 & 0.23 & \textbf{0.09} & \underline{0.02} & \textbf{0.11} & \textbf{0.21} \\
         \cmidrule(lr){3-6} \cmidrule(lr){7-10}
         & \textbf{RotatE} & 0.18 & \underline{0.14} & \underline{0.20} & 0.26 & 0.03 & 0.01 & 0.03 & 0.05 \\
         & \textbf{+\textsc{ImbalancE}} & \textbf{0.19} & \underline{0.14} & \underline{0.20} & \textbf{0.29} & 0.03 & 0.01 & 0.03 & 0.08 \\
         \midrule
         \multirow{4}{*}{\textbf{WN18RR}} & \textbf{ComplEx-N3} & 0.46 & 0.40 & 0.47 & 0.55 & 0.28 & 0.23 & 0.28 & 0.38 \\
         & \textbf{+\textsc{ImbalancE}} & 0.48 & 0.40 & 0.51 & \textbf{0.61} & 0.31 & 0.25 & 0.32 & \textbf{0.46} \\
         \cmidrule(lr){3-6} \cmidrule(lr){7-10}
         & \textbf{RotatE} & 0.49 & 0.45 & 0.50 & 0.57 & 0.32 & \underline{0.27} & 0.33 & 0.42 \\
         & \textbf{+\textsc{ImbalancE}} & \textbf{0.51} & \textbf{0.46} & \textbf{0.55} & 0.60 & \textbf{0.33} & \underline{0.27} & \textbf{0.35} & 0.44 \\
         \bottomrule
    \end{tabular}
    \label{tab:results}
\end{table*}

\subsubsection{Baseline Comparison} We compare our method with two different approaches in Table~\ref{tab:additional_baselines}. KG-Mixup~\cite{shomer2023degree_bias}
was designed to counter a vague notion of degree bias and to enhance performance on low-degree entities. However, we show how it provides little benefit on imbalanced triples compared to \textsc{ImbalancE}. CSProm-KG~\cite{chen2023cspromkg}, on the other hand, is a method that integrates an LLM to enhance link prediction and is way superior in aggregate metrics to ComplEx and RotatE ($MRR=0.36$, $H@10=0.54$ on the entire FB15k-237 test set). On imbalanced triples, \textsc{ImbalancE} outperforms it or falls behind slightly, despite being way more lightweight. Therefore, our approach makes simpler models perform on the same level as more involved ones, preserving the computational efficiency that CSProm-KG and other LLM-enhanced KGE models cannot offer.

\begin{table*}[t]
    \centering
    \caption{Results on FB15k-237 for \textsc{ImbalancE} and two baselines. The best value for each metric on each dataset is reported in \textbf{bold} except in case of a tie, when they are \underline{underlined}. \textsc{ImbalancE} outperforms previous work to address ``degree bias'', and it also brings simpler and less capable KGE models to the same levels of a much more computationally intensive option.}
    \begin{tabular}{llcccccccccc}
            \toprule
             \multirow{2}{*}{\textbf{Dataset}} & \multirow{2}{*}{\textbf{Model}} & \multicolumn{4}{c}{\textbf{High-Low}} & \multicolumn{4}{c}{\textbf{Low-High}}\\
             \cmidrule(lr){3-6} \cmidrule(lr){7-10}
             & & \textbf{MRR} & \textbf{H@1} & \textbf{H@3} & \textbf{H@10} & \textbf{MRR} & \textbf{H@1} & \textbf{H@3} & \textbf{H@10} \\
            \midrule
             \multirow{5}{*}{\textbf{FB15k-237}} & \textbf{ComplEx-N3 + \textsc{ImbalancE}} & \underline{0.13} & 0.05 & \textbf{0.14} & 0.28 & 0.08 & 0.04 & 0.08 & \underline{0.17} \\
             \cmidrule(lr){3-6} \cmidrule(lr){7-10}
             & \textbf{RotatE + \textsc{ImbalancE}} & \underline{0.13} & \textbf{0.06} & 0.11 & \textbf{0.36} & 0.08 & 0.04 & \underline{0.09} & \underline{0.17} \\
         \cmidrule(lr){2-10}
             & \textbf{TuckER}~\cite{balazevic2019tucker} & 0.08 & 0.03 & 0.08 & 0.21 & 0.08 & 0.04 & 0.07 & 0.16 \\
             & \textbf{+KG-Mixup}~\cite{shomer2023degree_bias} & 0.09 & 0.03 & 0.08 & 0.24 & 0.08 & 0.03 & 0.08 & \underline{0.17} \\
             \cmidrule(lr){2-10}
             & \textbf{CSProm-KG}~\cite{chen2023cspromkg} & 0.09 & 0.02 & 0.07 & 0.29 & \textbf{0.09} & \textbf{0.05} & \underline{0.09} & \underline{0.17} \\
            \bottomrule
    \end{tabular}
    \label{tab:additional_baselines}
\end{table*}

\subsubsection{Loss function terms contribution} To gauge the contribution of the training context term and of the oracle term, we run separate experiments where we switch off alternately one or the other. The results for FB15k-237 are reported in Table~\ref{tab:ablation_loss}, while those for WN18RR and Yago3-10 are in Appendix~\ref{apx:additional_experiments}. Numbers clearly show that both terms provide a positive contribution on their own, while their joint contribution outperforms the single terms, proving their complementarity. This supports both our claims on how grounding the representation of the anchor node in a query-friendly way is essential and on the importance of adjusting the embeddings of other entities. We explain the bigger impact of the oracle term as it directly contributes to the optimization of the anchor embedding. This also makes the gains more dependent on the quality of the oracle, making its selection a critical design choice.

\begin{table}[t]
    \centering
    \caption{Results on FB15k-237 isolating the two terms of the loss function. Both terms provide complementary contributions.}
    \footnotesize
      \begin{tabular}{llcccccc}
        \toprule
         \multirow{2}{*}{\textbf{Dataset}} & \multirow{2}{*}{\textbf{Model}} & \multicolumn{2}{c}{\textbf{High-Low}} & \multicolumn{2}{c}{\textbf{Low-High}}\\
         \cmidrule(lr){3-4} \cmidrule(lr){5-6}
         & & \textbf{MRR} & \textbf{H@10} & \textbf{MRR} &  \textbf{H@10} \\
        
        \midrule
        
         \multirow{8}{*}{\textbf{FB15k-237}} & \textbf{ComplEx-N3} & 0.03 & 0.06 & 0.03 & 0.08 \\
         & \textbf{+Context Only} & 0.05 & 0.14 & 0.05 & 0.12 \\
         & \textbf{+Oracle Only} & 0.10 & 0.22 & \underline{0.08} & \underline{0.17}  \\
         & \textbf{+\textsc{ImbalancE}} & \textbf{0.13} & \textbf{0.28} & \underline{0.08} & \underline{0.17} \\
         \cmidrule(lr){3-4} \cmidrule(lr){5-6}
         & \textbf{RotatE} & 0.02 & 0.05 & 0.04 & 0.09  \\
         & \textbf{+Context Only} & 0.06 & 0.16 & 0.06 & 0.14 \\
         & \textbf{+Oracle Only} & 0.12 & 0.34 & \underline{0.08} & \underline{0.17} \\
         & \textbf{+\textsc{ImbalancE}} & \textbf{0.13} & \textbf{0.36} & \underline{0.08} &  \underline{0.17} \\

         \bottomrule
    \end{tabular}
    
    \label{tab:ablation_loss}
\end{table}

\subsubsection{Time-Complexity and Latency}
The time-complexity of one epoch of \textsc{ImbalancE} is $\mathcal{O} \left( k \cdot (\lvert \mathcal{C}_q \rvert + \lvert \Omega_q \rvert ) \right)$. In fact, it applies the KGE scoring function and computes the gradients for the embeddings of the triples in $\mathcal{C}_q$ and $\Omega_q$. The already low time-complexity to train a traditional KGE for one epoch amounts to $\mathcal{O}(\lvert \mathcal{G} \rvert (\eta + 1) k)$, where $\mathcal{G}$ is the training graph and $\eta$ the number of negatives. Instead of re-traning such model, applying \textsc{ImbalancE} to a test set $\mathcal{T}$ is more efficient, as the size of $\mathcal{T}$ is typically orders of magnitude smaller than $\mathcal{G}$. Moreover, $\mathcal{C}_q$ is a limited subset of $\mathcal{G}$ and the number of oracle triples are only a few dozens at most (in our experiments always $\leq 50$), yielding {\small$\lvert \mathcal{T} \rvert \cdot (\lvert \mathcal{C} \rvert + \lvert \Omega \rvert ) \ll \lvert \mathcal{G} \rvert$}. Finally, \textsc{ImbalancE} requires fewer epochs to reach convergence, thus confirming its efficiency even compared to the strong baseline. The run time latency of the system is reported in Table~\ref{tab:latency}. The extreme efficiency of $\textsc{ImbalancE}$ is not decreased by the pre-processing step to extract the oracle triples. In fact, in our experiments, this time amounted to 63sec for FB15k-237, 79sec for WN18RR, and 372sec for Yago3-10.

\begin{table}[h!]
\centering
\footnotesize
\caption{Latency of \textsc{ImbalancE} compared to re-training the KGE model from scratch. Runtime is reported in seconds and 30 epochs of latent search are considered, though convergence is typically reached within the first 10.}
\begin{tabular}{cccc}
     % \cmidrule(lr){2-4}
     \toprule
      & \textbf{Model} & \textbf{FB15k-237} & \textbf{Yago3-10} \\
     \midrule
     Single query & ~\textsc{ImbalancE} & 0.91 & 1.39 \\
     Full test set & ~\textsc{ImbalancE} & 440.7 & 400.1 \\
     Pre-training & RotatE & 935.9 & \num{24131.3} \\
     \bottomrule
\end{tabular}

\label{tab:latency}
\end{table}

\section{Limitations and Future Directions}
\label{sec:limitations}

If this work sheds light on a key issue, it has limitations that we will address in future work. First, \textsc{ImbalancE} can only be applied on queries for which the training context is non-empty. Overcoming this limitation would mean refining the selection of training triples \textit{tightly related} to the query. This would be of great interest, as it would explain which triples impact a prediction the most.
Second, the benefit of $\textsc{ImbalancE}$ is dataset-dependent and might be tamed by less polarized degree distributions (e.g., WN18RR) or weaker oracle inputs. As Yago3-10 showed, the quality of the triples generated by the oracle affects the quality of the results, which makes our results dependent on the LLM (and the oracle more in general). We leave for future work an in-depth study of different oracles, that go beyond LLMs and potentially include human feedback.

\section{Conclusions}
\label{sec:conclusions}

This work has introduced the degree imbalance problem, that heavily affects a wide variety of KGE models and hinders their predictive power. We provided deep insights on the learning issue from which the problem stems, giving evidence consistent with two distinct failure modes for low-degree entities — overfitting on one side and failed convergence on the other. To mitigate the issue, we proposed \textsc{ImbalancE}, the first inference-time latent search method applied in the realm of KGE models. Its efficacy on highly imbalanced triples and its efficiency open the door to the reliable application of link prediction in use cases where degree imbalance is a consistent concern.

\paragraph*{Supplemental Material Statement:} the paper contains comprehensive experimental details to support reproducibility, including instructions to determine dataset splits and their statistics, and  all relevant hyperparameters used throughout the study. The codebase to reproduce the results is available at \url{https://github.com/Accenture/AmpliGraph/tree/paper/ISWC2026_ImbalancE}.

\newpage

\bibliographystyle{splncs04}
\bibliography{imbalance.bib}

\begin{thebibliography}{10}
\providecommand{\url}[1]{\texttt{#1}}
\providecommand{\urlprefix}{URL }
\providecommand{\doi}[1]{https://doi.org/#1}

\bibitem{balazevic2019tucker}
Balazevic, I., Allen, C., Hospedales, T.M.: Tucker: Tensor factorization for knowledge graph completion. In: Inui, K., Jiang, J., Ng, V., Wan, X. (eds.) Proceedings of the 2019 Conference on Empirical Methods in Natural Language Processing and the 9th International Joint Conference on Natural Language Processing, {EMNLP-IJCNLP} 2019, Hong Kong, China, November 3-7, 2019. pp. 5184--5193. Association for Computational Linguistics (2019). \doi{10.18653/V1/D19-1522}, \url{https://doi.org/10.18653/v1/D19-1522}

\bibitem{bonnet2024searchinglatentprogramspaces}
Bonnet, C., Macfarlane, M.V.: Searching latent program spaces (2024), \url{https://arxiv.org/abs/2411.08706}

\bibitem{bordes2013transe}
Bordes, A., Usunier, N., Garcia-Duran, A., Weston, J., Yakhnenko, O.: Translating embeddings for modeling multi-relational data. In: Burges, C., Bottou, L., Welling, M., Ghahramani, Z., Weinberger, K. (eds.) Advances in Neural Information Processing Systems. vol.~26. Curran Associates, Inc. (2013)

\bibitem{cao2024survey_kge}
Cao, J., Fang, J., Meng, Z., Liang, S.: Knowledge graph embedding: A survey from the perspective of representation spaces. ACM Comput. Surv.  \textbf{56}(6) (mar 2024). \doi{10.1145/3643806}, \url{https://doi.org/10.1145/3643806}

\bibitem{chen2023cspromkg}
Chen, C., Wang, Y., Sun, A., Li, B., Lam, K.: Dipping plms sauce: Bridging structure and text for effective knowledge graph completion via conditional soft prompting. In: Rogers, A., Boyd{-}Graber, J.L., Okazaki, N. (eds.) Findings of the Association for Computational Linguistics: {ACL} 2023, Toronto, Canada, July 9-14, 2023. pp. 11489--11503. Association for Computational Linguistics (2023). \doi{10.18653/V1/2023.FINDINGS-ACL.729}, \url{https://doi.org/10.18653/v1/2023.findings-acl.729}

\bibitem{dettmetters2018conve}
Dettmers, T., Minervini, P., Stenetorp, P., Riedel, S.: Convolutional 2d knowledge graph embeddings. In: McIlraith, S.A., Weinberger, K.Q. (eds.) Proceedings of the Thirty-Second {AAAI} Conference on Artificial Intelligence, (AAAI-18), the 30th innovative Applications of Artificial Intelligence (IAAI-18), and the 8th {AAAI} Symposium on Educational Advances in Artificial Intelligence (EAAI-18), New Orleans, Louisiana, USA, February 2-7, 2018. pp. 1811--1818. {AAAI} Press (2018). \doi{10.1609/AAAI.V32I1.11573}, \url{https://doi.org/10.1609/aaai.v32i1.11573}

\bibitem{dong2014knowledge_vault}
Dong, X., Gabrilovich, E., Heitz, G., Horn, W., Lao, N., Murphy, K., Strohmann, T., Sun, S., Zhang, W.: Knowledge vault: a web-scale approach to probabilistic knowledge fusion. In: Proceedings of the 20th ACM SIGKDD International Conference on Knowledge Discovery and Data Mining. p. 601–610. KDD '14, Association for Computing Machinery, New York, NY, USA (2014). \doi{10.1145/2623330.2623623}, \url{https://doi.org/10.1145/2623330.2623623}

\bibitem{he2015kg2e}
He, S., Liu, K., Ji, G., Zhao, J.: Learning to represent knowledge graphs with gaussian embedding. In: Proceedings of the 24th ACM International on Conference on Information and Knowledge Management. p. 623–632. CIKM '15, Association for Computing Machinery, New York, NY, USA (2015). \doi{10.1145/2806416.2806502}, \url{https://doi.org/10.1145/2806416.2806502}

\bibitem{hogan2021survey}
Hogan, A., Blomqvist, E., Cochez, M., D’amato, C., Melo, G.D., Gutierrez, C., Kirrane, S., Gayo, J.E.L., Navigli, R., Neumaier, S., Ngomo, A.C.N., Polleres, A., Rashid, S.M., Rula, A., Schmelzeisen, L., Sequeda, J., Staab, S., Zimmermann, A.: Knowledge graphs  \textbf{54}(4) (jul 2021). \doi{10.1145/3447772}, \url{https://doi.org/10.1145/3447772}

\bibitem{ji2015transd}
Ji, G., He, S., Xu, L., Liu, K., Zhao, J.: Knowledge graph embedding via dynamic mapping matrix. In: Zong, C., Strube, M. (eds.) Proceedings of the 53rd Annual Meeting of the Association for Computational Linguistics and the 7th International Joint Conference on Natural Language Processing (Volume 1: Long Papers). pp. 687--696. Association for Computational Linguistics, Beijing, China (Jul 2015). \doi{10.3115/v1/P15-1067}, \url{https://aclanthology.org/P15-1067/}

\bibitem{kamigaito2022survey}
Kamigaito, H., Hayashi, K.: Comprehensive analysis of negative sampling in knowledge graph representation learning. In: Chaudhuri, K., Jegelka, S., Song, L., Szepesv{\'{a}}ri, C., Niu, G., Sabato, S. (eds.) International Conference on Machine Learning, {ICML} 2022, 17-23 July 2022, Baltimore, Maryland, {USA}. Proceedings of Machine Learning Research, vol.~162, pp. 10661--10675. {PMLR} (2022), \url{https://proceedings.mlr.press/v162/kamigaito22a.html}

\bibitem{bengio2015adam}
Kingma, D.P., Ba, J.: Adam: {A} method for stochastic optimization. In: Bengio, Y., LeCun, Y. (eds.) 3rd International Conference on Learning Representations, {ICLR} 2015, San Diego, CA, USA, May 7-9, 2015, Conference Track Proceedings (2015), \url{http://arxiv.org/abs/1412.6980}

\bibitem{lacroix2018complex_n3}
Lacroix, T., Usunier, N., Obozinski, G.: Canonical tensor decomposition for knowledge base completion. In: International Conference on Machine Learning. pp. 2869--2878 (2018)

\bibitem{lin2015transr}
Lin, Y., Liu, Z., Sun, M., Liu, Y., Zhu, X.: Learning entity and relation embeddings for knowledge graph completion. p. 2181–2187. AAAI'15, AAAI Press (2015)

\bibitem{madushanka2024survey}
Madushanka, T., Ichise, R.: Negative sampling in knowledge graph representation learning: {A} review. CoRR  \textbf{abs/2402.19195} (2024). \doi{10.48550/ARXIV.2402.19195}, \url{https://doi.org/10.48550/arXiv.2402.19195}

\bibitem{mahdisoltani2015yago3}
Mahdisoltani, F., Biega, J., Suchanek, F.M.: {YAGO3:} {A} knowledge base from multilingual wikipedias. In: Seventh Biennial Conference on Innovative Data Systems Research, {CIDR} 2015, Asilomar, CA, USA, January 4-7, 2015, Online Proceedings. www.cidrdb.org (2015), \url{http://cidrdb.org/cidr2015/Papers/CIDR15\_Paper1.pdf}

\bibitem{mohamed2020popularity_bias}
Mohamed, A., Parambath, S., Kaoudi, Z., Aboulnaga, A.: Popularity agnostic evaluation of knowledge graph embeddings. In: Peters, J., Sontag, D. (eds.) Proceedings of the 36th Conference on Uncertainty in Artificial Intelligence (UAI). Proceedings of Machine Learning Research, vol.~124, pp. 1059--1068. PMLR (03--06 Aug 2020), \url{https://proceedings.mlr.press/v124/mohamed20a.html}

\bibitem{nickel2012factorizing}
Nickel, M., Tresp, V., Kriegel, H.: Factorizing {YAGO:} scalable machine learning for linked data. In: Mille, A., Gandon, F., Misselis, J., Rabinovich, M., Staab, S. (eds.) Proceedings of the 21st World Wide Web Conference 2012, {WWW} 2012, Lyon, France, April 16-20, 2012. pp. 271--280. {ACM} (2012). \doi{10.1145/2187836.2187874}, \url{https://doi.org/10.1145/2187836.2187874}

\bibitem{rossi2021relation_bias}
Rossi, A., Barbosa, D., Firmani, D., Matinata, A., Merialdo, P.: Knowledge graph embedding for link prediction: A comparative analysis. ACM Trans. Knowl. Discov. Data  \textbf{15}(2) (Jan 2021). \doi{10.1145/3424672}, \url{https://doi.org/10.1145/3424672}

\bibitem{sardina2024surveyknowledgegraphstructure}
Sardina, J., Kelleher, J.D., O'Sullivan, D.: A survey on knowledge graph structure and knowledge graph embeddings (2024), \url{https://arxiv.org/abs/2412.10092}

\bibitem{shomer2023degree_bias}
Shomer, H., Jin, W., Wang, W., Tang, J.: Toward degree bias in embedding-based knowledge graph completion. In: Proceedings of the ACM Web Conference 2023. p. 705–715. WWW '23, Association for Computing Machinery, New York, NY, USA (2023). \doi{10.1145/3543507.3583544}, \url{https://doi.org/10.1145/3543507.3583544}

\bibitem{sun2019rotate}
Sun, Z., Deng, Z., Nie, J., Tang, J.: Rotate: Knowledge graph embedding by relational rotation in complex space. In: 7th International Conference on Learning Representations, {ICLR} 2019, New Orleans, LA, USA, May 6-9, 2019. OpenReview.net (2019), \url{https://openreview.net/forum?id=HkgEQnRqYQ}

\bibitem{toutanova2015fb15k237}
Toutanova, K., Chen, D.: Observed versus latent features for knowledge base and text inference. In: Allauzen, A., Grefenstette, E., Hermann, K.M., Larochelle, H., Yih, S.W. (eds.) Proceedings of the 3rd Workshop on Continuous Vector Space Models and their Compositionality, {CVSC} 2015, Beijing, China, July 26-31, 2015. pp. 57--66. Association for Computational Linguistics (2015). \doi{10.18653/V1/W15-4007}, \url{https://doi.org/10.18653/v1/W15-4007}

\bibitem{trouillon2016complex}
Trouillon, T., Welbl, J., Riedel, S., Gaussier, {\'E}., Bouchard, G.: Complex embeddings for simple link prediction. In: ICML. pp. 2071--2080 (2016)

\bibitem{wang2014transh}
Wang, Z., Zhang, J., Feng, J., Chen, Z.: Knowledge graph embedding by translating on hyperplanes. In: Proceedings of the Twenty-Eighth AAAI Conference on Artificial Intelligence. p. 1112–1119. AAAI'14, AAAI Press (2014)

\bibitem{yang2015distmult}
Yang, B., Yih, W., He, X., Gao, J., Deng, L.: Embedding entities and relations for learning and inference in knowledge bases. In: Bengio, Y., LeCun, Y. (eds.) 3rd International Conference on Learning Representations, {ICLR} 2015, San Diego, CA, USA, May 7-9, 2015, Conference Track Proceedings (2015), \url{http://arxiv.org/abs/1412.6575}

\bibitem{zhang2016kg_for_recommender}
Zhang, F., Yuan, N.J., Lian, D., Xie, X., Ma, W.Y.: Collaborative knowledge base embedding for recommender systems. In: Proceedings of the 22nd ACM SIGKDD International Conference on Knowledge Discovery and Data Mining. p. 353–362. KDD '16, Association for Computing Machinery, New York, NY, USA (2016). \doi{10.1145/2939672.2939673}, \url{https://doi.org/10.1145/2939672.2939673}

\bibitem{zhu2021neural}
Zhu, Z., Zhang, Z., Xhonneux, L.P., Tang, J.: Neural bellman-ford networks: A general graph neural network framework for link prediction. Advances in Neural Information Processing Systems  \textbf{34} (2021)

\end{thebibliography}

\clearpage

% -------- Appendices ------

\appendix

% \section{Validation-based Latent Search Early Stopping}

\section{Degree Imbalance for Additional Models}
\label{apx:degree_imbalance_more_model}

\begin{figure}[tbh]
    \centering
    % --- FB15k-237 ---
    \begin{subfigure}{\linewidth}
        \centering
        \includegraphics[width=0.7\linewidth]{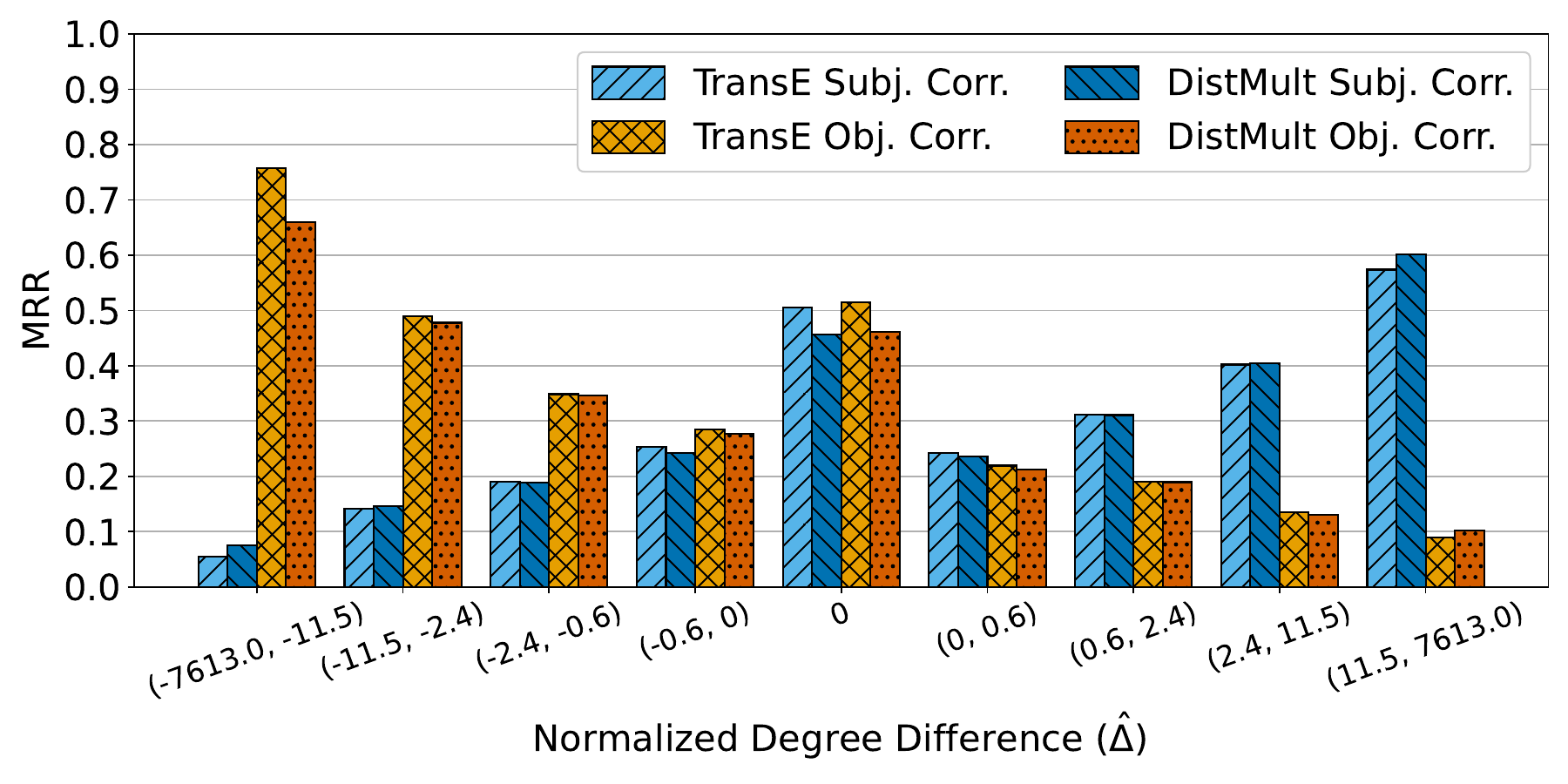}
        \caption{FB15k-237}
        \label{fig:mrr_fb15k237_transe_distmult}
    \end{subfigure}
    % --- YAGO3-10 ---
    \begin{subfigure}{\linewidth}
        \centering
        \includegraphics[width=0.7\linewidth]{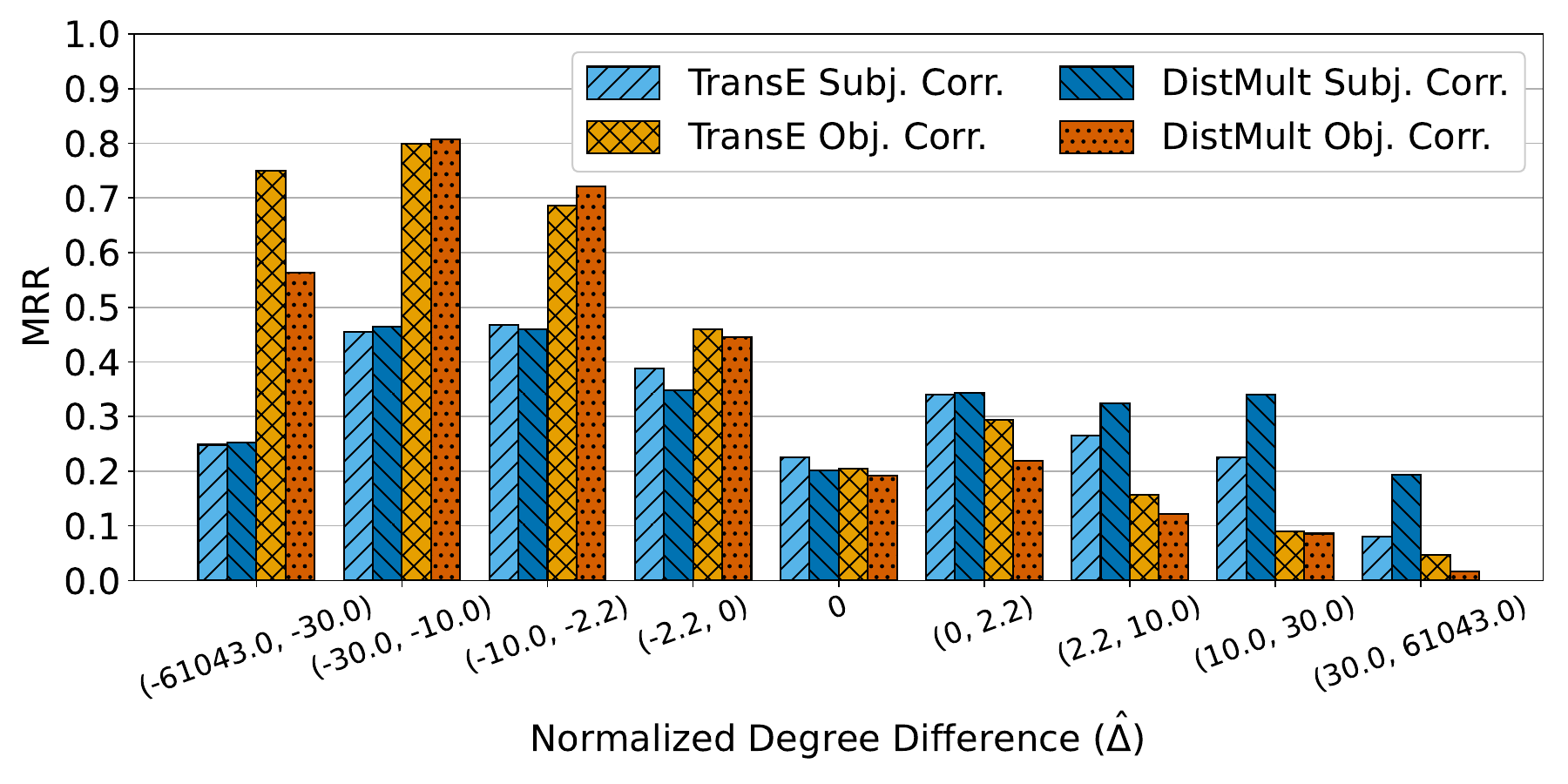}
        \caption{YAGO3-10}
        \label{fig:mrr_yago310_transe_distmult}
    \end{subfigure}
    \caption{Performance of conventional KGE models on test triples binned by normalized degree difference. Bins are obtained using quartiles of $\lvert \hat{\Delta} (t) \rvert$. On the left hand side we have triples with high-degree object and low-degree subject, on the right hand side, high-degree subjects and low-degree objects. We isolate subject and object corruption.}
    \label{fig:mrr_degree_diff_transe_distmult}
\end{figure}

\begin{figure}[tbh]
    \centering
    \includegraphics[width=\linewidth]{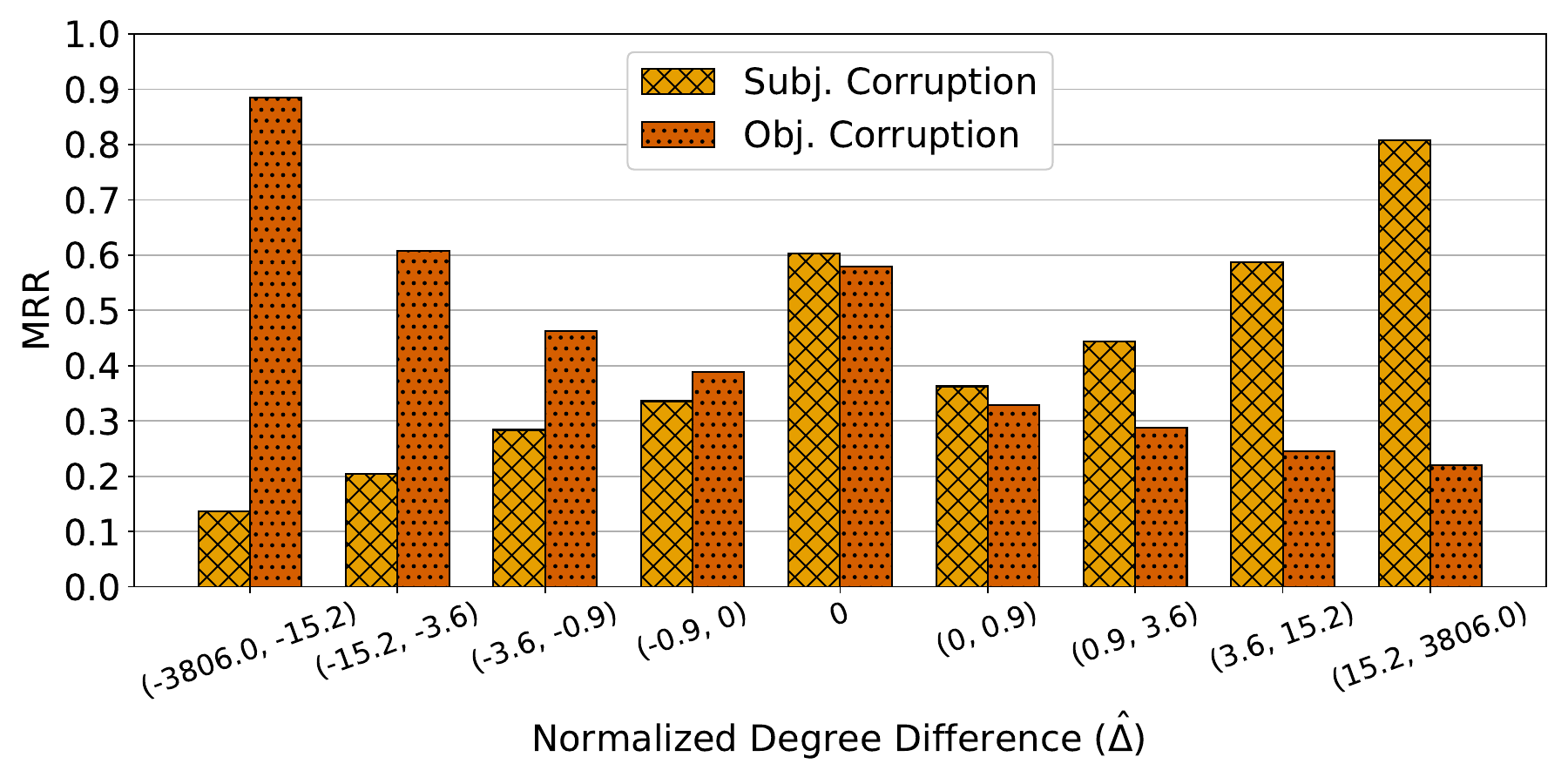}
    \caption{Performance of NBFNet on FB15k-237 test triples binned by normalized degree difference. Bins are obtained using quartiles of $\lvert \hat{\Delta} \rvert$. On the left hand side we have triples with high-degree object and low-degree subject, on the right hand side, high-degree subjects and low-degree objects. We isolate subject and object corruption.}
    \label{fig:nbfnet_deg_imbalance}
\end{figure}

In Table~\ref{tab:transe_distmult_perf} we report the aggregate metrics for TransE \cite{bordes2013transe}, DistMult \cite{yang2015distmult} and NBFNet \cite{zhu2021neural}, while in Figures~\ref{fig:mrr_degree_diff_transe_distmult} and \ref{fig:nbfnet_deg_imbalance} we plot the Mean Reciprocal Rank as a function of the degree imbalance for TransE, DistMult, and NBFNet. As observed in the main text of the paper, all model predictions are heavily affected by the degree imbalance. In fact, the gap in performance between the high-degree and low-degree entity predictions increases as the degree imbalance increases. Notably, even NBFNet, that achieves superior aggregate performance, struggles in the same way.

As an additional note, we remark that NBFNet cannot benefit from the application of \textsc{ImbalancE}. In fact, the representations of different entities are heavily entangled within the message-passing architecture. This prevents the independent update of the anchor and of the other target entities, while keeping frozen the relations and all the other embeddings, making the application of our approach not scalable and prone to overfitting.

\begin{table}[h]
\footnotesize
\caption{Aggregate performance of KGE models on FB15k-237 and YAGO3-10}
\centering
\begin{tabular}{c c c c c}
\toprule
 & \multicolumn{2}{c}{\textbf{FB15k-237}} & \multicolumn{2}{c}{\textbf{YAGO3-10}} \\
\cmidrule(lr){2-3} \cmidrule(lr){4-5}
 & \textbf{MRR} & \textbf{H@10} & \textbf{MRR} & \textbf{H@10} \\
\midrule
TransE    & 0.31 & 0.49 & 0.35 & 0.55 \\
DistMult  & 0.30 & 0.48 & 0.34 & 0.53 \\
NBFNet & 0.41 & 0.60 & - & - \\
\bottomrule
\end{tabular}
\label{tab:transe_distmult_perf}
\end{table}

\section{Dataset Statistics}
\label{apx:dataset_stats}
Statistics of the three datasets considered in our experiments are reported in Table~\ref{tab:stats_datasets}.

\begin{table}[h!]
    \centering
    \footnotesize
    \caption{Statistics of the datasets.}
    \begin{tabular}{lccccc}
        \toprule
         \textbf{Dataset} & \textbf{\#Entities} & \textbf{\#Relations} & \textbf{\#Train} & \textbf{\#Valid} & \textbf{\#Test} \\
         \midrule
         FB15k-237 & \num{15541} & \num{237} & \num{272115} & \num{17535} & \num{20466} \\
         WN18RR & \num{40943} & \num{18} & \num{86835} & \num{3034} & \num{3134} \\
         Yago3-10 & \num{123182} & \num{37} & \num{1079040} & \num{5000} & \num{5000} \\
         \bottomrule
    \end{tabular}
    \label{tab:stats_datasets}
\end{table}

\section{Oracle Technical Details}
\label{apx:oracle}
For all our experiments, we have used \texttt{NovaSearch/stella\_en\_400M\_v5}\footnote{\url{https://huggingface.co/NovaSearch/stella_en_400M_v5}} with an embedding dimension of 1,024 to encode the labels and descriptions of the entities in our datasets (with the exception of Yago3-10, for which we used only the labels). Given $q = (s, p, ?)$, to generate the $m$ triples in $\Omega_q$, we have extracted the top-$m$ entities closest to the barycenter of the set $\{ e_j \in \mathcal{G} | \{ (s, p, e_j) \in \mathcal{C}_q \}$ in the embedding space of the LLM. The distance between entities was computed using the cosine similarity.

As querying an LLM can introduce a significant overhead, we have extracted the LLM-triples as a preprocessing step ahead of running the experiments. In this way, the method remains lightweight.

\section{Additional Experiments}
\label{apx:additional_experiments}

\subsubsection{Loss Function Terms Contribution} As Table~\ref{tab:ablation_loss_others} shows, the two terms in conjunction give the best results. In Yago3-10 and WN18RR, the contribution of single terms is more limited compared to FB15k-237 as a consequence of the overall more limited impact of \textsc{ImbalancE} on these two datasets, as shown and justified in Section~\ref{sec:results}.

\begin{table*}[h]
    \centering
    \caption{Results on WN18RR, and YAGO3-10 isolating the contribution of the two terms of the loss function.}
    \footnotesize
      \begin{tabular}{llcccccccccc}
        \toprule
         \multirow{2}{*}{\textbf{Dataset}} & \multirow{2}{*}{\textbf{Model}} & \multicolumn{4}{c}{\textbf{High-Low}} & \multicolumn{4}{c}{\textbf{Low-High}}\\
         \cmidrule(lr){3-6} \cmidrule(lr){7-10}
         & & \textbf{MRR} & \textbf{H@1} & \textbf{H@3} & \textbf{H@10} & \textbf{MRR} & \textbf{H@1} & \textbf{H@3} & \textbf{H@10} \\
        
        \midrule

        % \multirow{8}{*}{\textbf{FB15k-237}} & \textbf{ComplEx-N3} & 0.03 & 0.01 & 0.01 & 0.06 & 0.03 & 0.01 & 0.02 & 0.08 \\
        %  & \textbf{+Context Only} & 0.05 & 0.01 & 0.04 & 0.14 & 0.05 & 0.02 & 0.04 & 0.12 \\
        %  & \textbf{+Oracle Only} & 0.10 & 0.05 & 0.09 & 0.22 & 0.08 & 0.03 & 0.07 & \underline{0.17}  \\
        %  & \textbf{+\textsc{ImbalancE}} & \underline{0.13} & 0.05 & \textbf{0.14} & 0.28 & \underline{0.08} & \underline{0.04} & 0.08 & \underline{0.17} \\
        %  \cmidrule(lr){3-6} \cmidrule(lr){7-10}
        %  & \textbf{RotatE} & 0.02 & 0.01 & 0.01 & 0.05 & 0.04 & 0.01 & 0.04 & 0.09  \\
        %  & \textbf{+Context Only} & 0.06 & 0.02 & 0.06 & 0.16 & 0.06 & 0.02 & 0.06 & 0.14 \\
        %  & \textbf{+Oracle Only} & 0.12 & 0.05 & 0.10 & 0.34 & \underline{0.08} & \underline{0.04} & \underline{0.09} & \underline{0.17} \\
        %  & \textbf{+\textsc{ImbalancE}} & \underline{0.13} & \textbf{0.06} & 0.11 & \textbf{0.36} & \underline{0.08} & \underline{0.04} & \underline{0.09} & \underline{0.17} \\

        %  \midrule

        \multirow{8}{*}{\textbf{Yago3-10}} & \textbf{ComplEx-N3} & 0.06 & 0.00 & 0.03 & 0.23 & 0.07 & \underline{0.02} & 0.07 & 0.17 \\
         & \textbf{+Context Only} & 0.06 & 0.00 & 0.03 & 0.23 & 0.06 & 0.02 & 0.07 & 0.17 \\
         & \textbf{+Oracle Only} & 0.06 & 0.00 & 0.03 & 0.23 & 0.07 & 0.00 & 0.10 & 0.20  \\
         & \textbf{+\textsc{ImbalancE}} & 0.06 & 0.00 & 0.03 & 0.23 & \textbf{0.08} & \underline{0.02} & \textbf{0.11} & \textbf{0.20} \\
         \cmidrule(lr){3-6} \cmidrule(lr){7-10}
         & \textbf{RotatE} & \underline{0.18} & \underline{0.14} & \underline{0.20} & 0.26 & 0.03 & 0.01 & 0.03 & 0.05  \\
         & \textbf{+Context Only} & 0.18 & 0.14 & 0.20 & 0.26 & 0.04 & 0.02 & 0.03 & 0.08\\
         & \textbf{+Oracle Only} & 0.19 & 0.14 & 0.20 & 0.29 & 0.03 & 0.01 & 0.03 & 0.06 \\
         & \textbf{+\textsc{ImbalancE}} & \textbf{0.19} & \underline{0.14} & \underline{0.20} & \textbf{0.29} & 0.03 & 0.01 & 0.03 & 0.08 \\
         
         \midrule
         
         \multirow{8}{*}{\textbf{WN18RR}} & \textbf{ComplEx-N3} & 0.46 & 0.40 & 0.47 & 0.55 & 0.28 & 0.23 & 0.28 & 0.38 \\
         & \textbf{+Context Only} & 0.45 & 0.40 & 0.47 & 0.55 & 0.28 & 0.22 & 0.28 & 0.38 \\
         & \textbf{+Oracle Only} & 0.48 & 0.40 & 0.51 & \underline{0.61} & 0.31 & 0.25 & 0.33 & 0.45  \\
         & \textbf{+\textsc{ImbalancE}} & 0.48 & 0.40 & 0.51 & \underline{0.61} & 0.31 & 0.25 & 0.32 & \textbf{0.46} \\
         \cmidrule(lr){3-6} \cmidrule(lr){7-10}
         & \textbf{RotatE} & 0.49 & 0.45 & 0.50 & 0.57 & 0.32 & \underline{0.27} & 0.33 & 0.42 \\
         & \textbf{+Context Only} & 0.49 & 0.45 & 0.5 & 0.58 & 0.32 & \underline{0.27} & 0.33 & 0.42 \\
         & \textbf{+Oracle Only} & 0.49 & 0.43 & 0.53 & 0.6 & 0.32 & \underline{0.27} & 0.34 & 0.42 \\
         & \textbf{+\textsc{ImbalancE}} & \textbf{0.51} & \textbf{0.46} & \textbf{0.55} & 0.60 & \textbf{0.33} & \underline{0.27} & \textbf{0.35} & 0.44 \\
         \bottomrule
    \end{tabular}
    \label{tab:ablation_loss_others}
\end{table*}

\subsubsection{Sankey Plots}
\begin{figure*}[h!]
    \centering
    \begin{subfigure}[b]{0.45\linewidth}
        \centering
        \includegraphics[width=\linewidth]{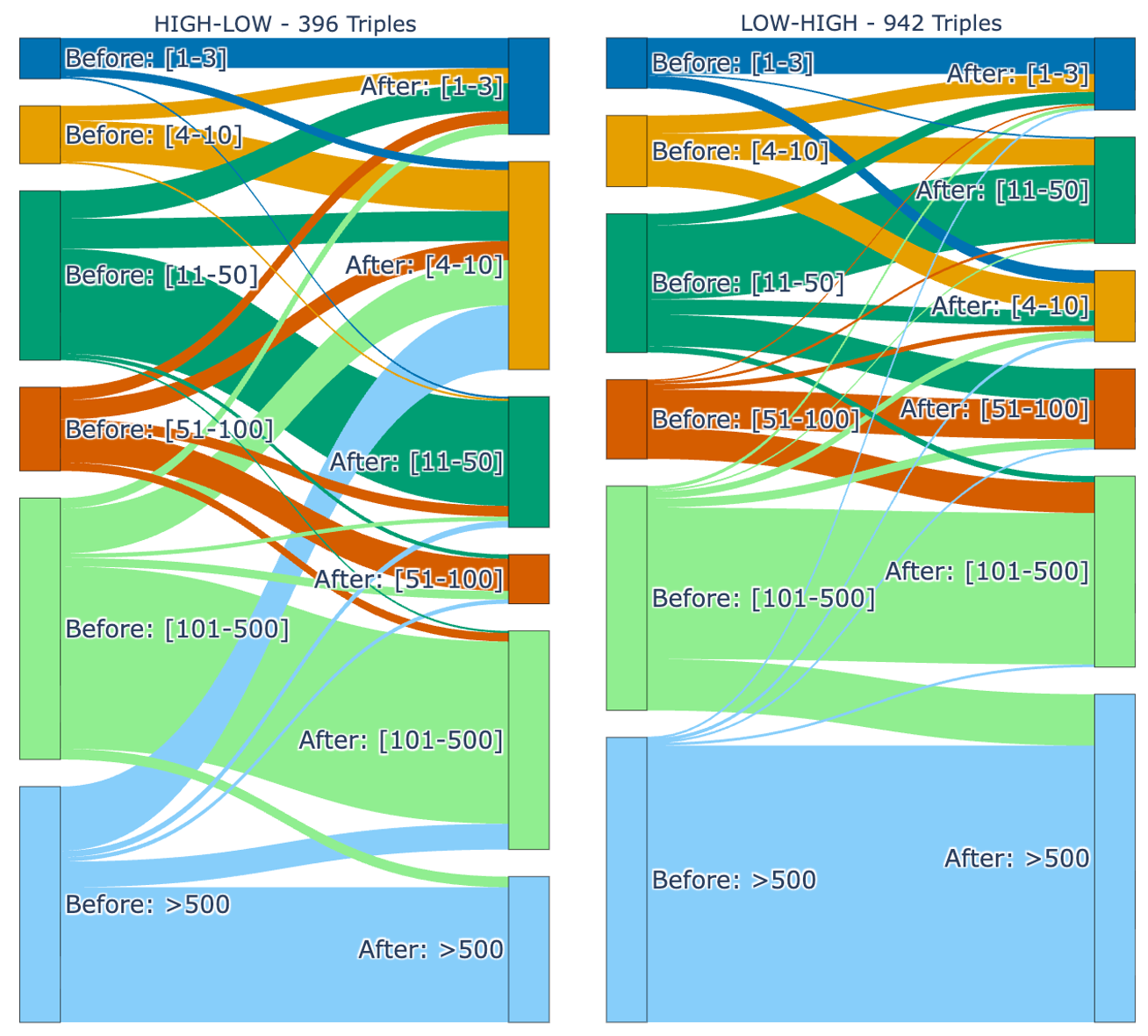}
        \caption{RotatE}
        \label{fig:rotate_sankey_fb15k-237}
    \end{subfigure}
    \hfill
    \begin{subfigure}[b]{0.45\linewidth}
        \centering
        \includegraphics[width=\linewidth]{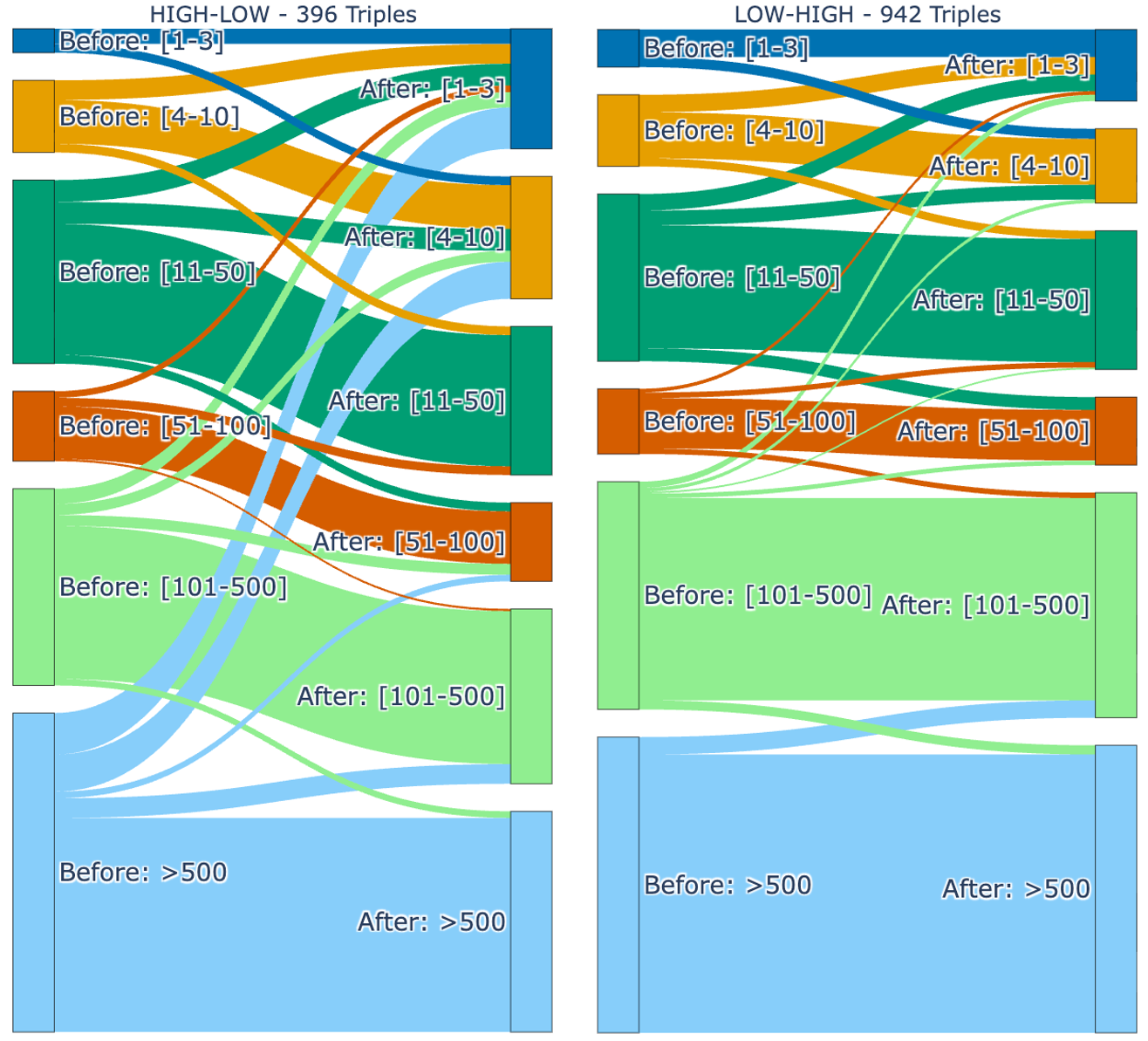}
        \caption{ComplEx}
        \label{fig:complex_sankey_fb15k-237}
    \end{subfigure}
    \caption{Sankey plots detailing how \textsc{ImbalancE} altered RotatE and ComplEx ranks on FB15k-237.}
    \label{fig:sankey_comparison_fb15k-237}
\end{figure*}

\begin{figure*}[h!]
    \centering
    \begin{subfigure}[b]{0.45\linewidth}
        \centering
        \includegraphics[width=\linewidth]{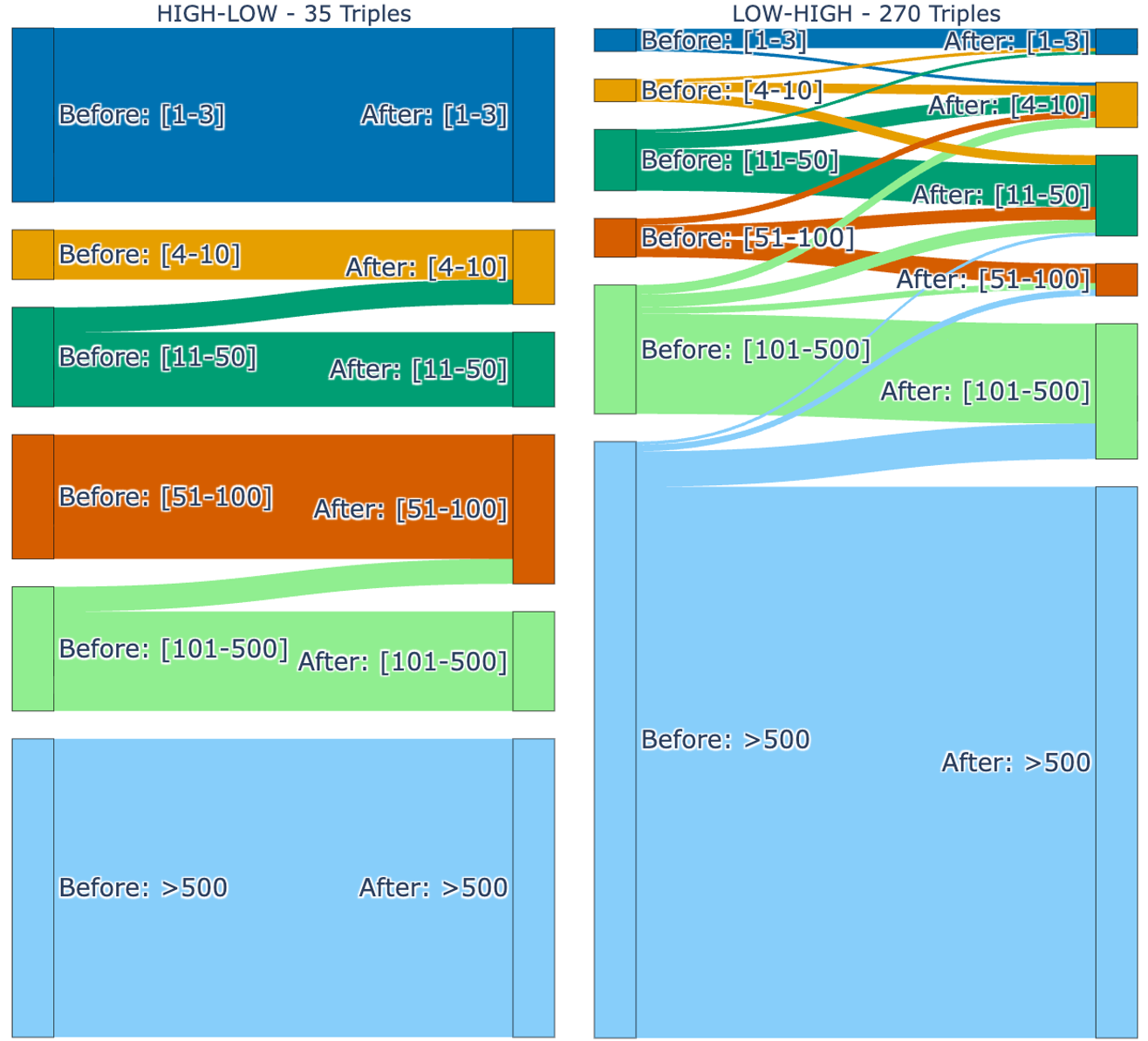}
        \caption{RotatE}
        \label{fig:rotate_sankey_yago310}
    \end{subfigure}
    \hfill
    \begin{subfigure}[b]{0.45\linewidth}
        \centering
        \includegraphics[width=\linewidth]{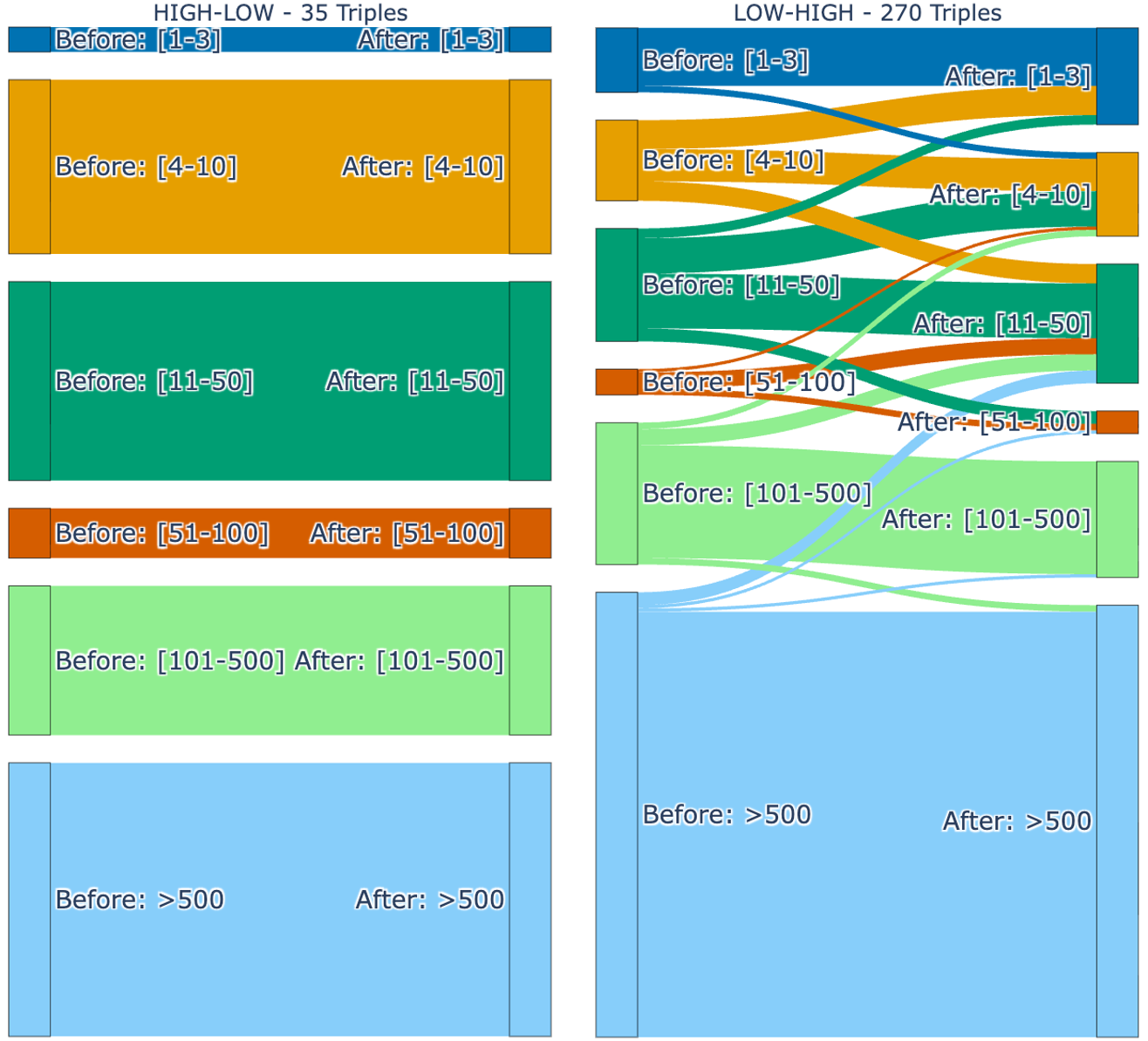}
        \caption{ComplEx}
        \label{fig:complex_sankey_yago310}
    \end{subfigure}
    \caption{Sankey plots detailing how \textsc{ImbalancE} altered RotatE and ComplEx ranks on Yago3-10.}
    \label{fig:sankey_comparison_yago310}
\end{figure*}

\begin{figure*}[h!]
    \centering
    \begin{subfigure}[b]{0.45\linewidth}
        \centering
        \includegraphics[width=\linewidth]{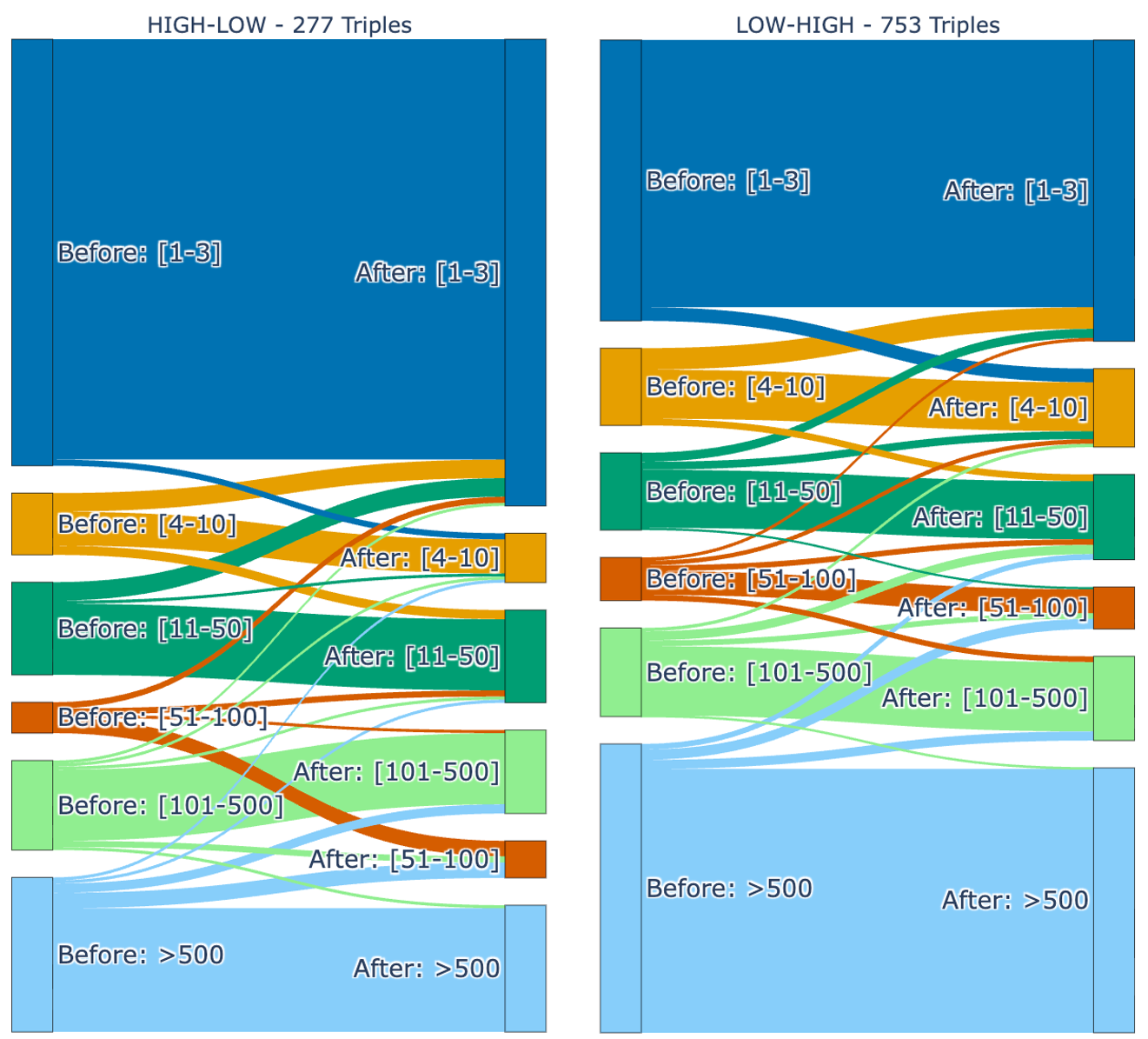}
        \caption{RotatE}
        \label{fig:rotate_sankey_wn18rr}
    \end{subfigure}
    \hfill
    \begin{subfigure}[b]{0.45\linewidth}
        \centering
        \includegraphics[width=\linewidth]{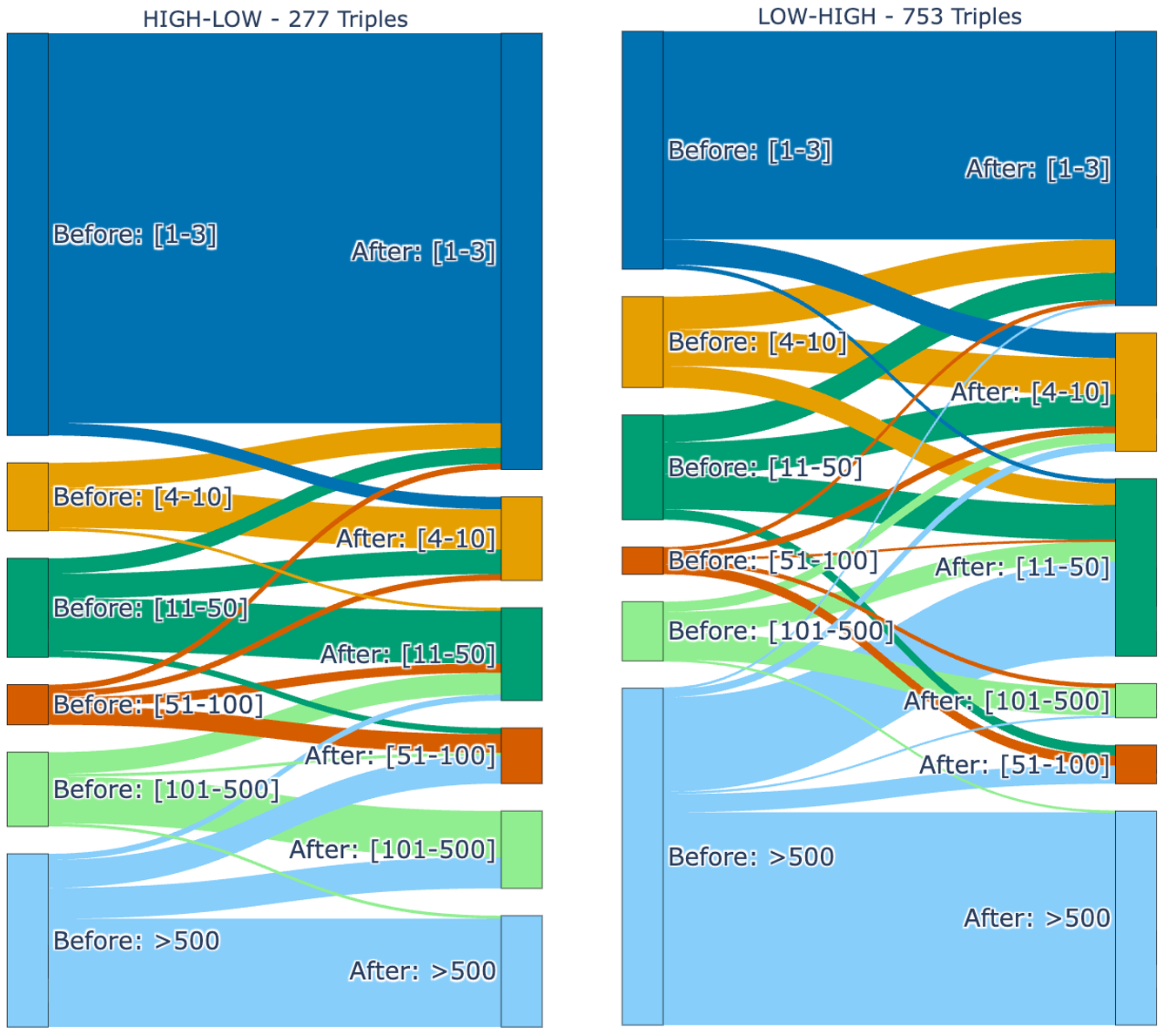}
        \caption{ComplEx}
        \label{fig:complex_sankey_wn18rr}
    \end{subfigure}
    \caption{Sankey plots detailing how \textsc{ImbalancE} altered RotatE and ComplEx ranks on WN18RR.}
    \label{fig:sankey_comparison_wn18rr}
\end{figure*}

To further explore how \textsc{ImbalancE} impacts the ranks assigned to test triples, we leverage Sankey plots that show how ranks flow from the values assigned by the pre-trained model (on the left), to the ranks assigned using \textsc{ImbalancE} (on the right). From Figures~\ref{fig:sankey_comparison_fb15k-237}-\ref{fig:sankey_comparison_yago310}-\ref{fig:sankey_comparison_wn18rr} we can observe how \textsc{ImbalancE} improves ranks of hundreds of positions. Although the metrics observed in Table~\ref{tab:results} do not necessarily show it, bringing up ranks that are $>500$ at the top positions of the ranking is a huge achievement, testifying to the value of the approach.

\subsubsection{Degenerate Behaviour Loss Function} 
\begin{figure*}[h]
    \centering
    
    % --- RotatE ---
    \begin{subfigure}{0.48\linewidth}
        \centering
        \includegraphics[width=\linewidth]{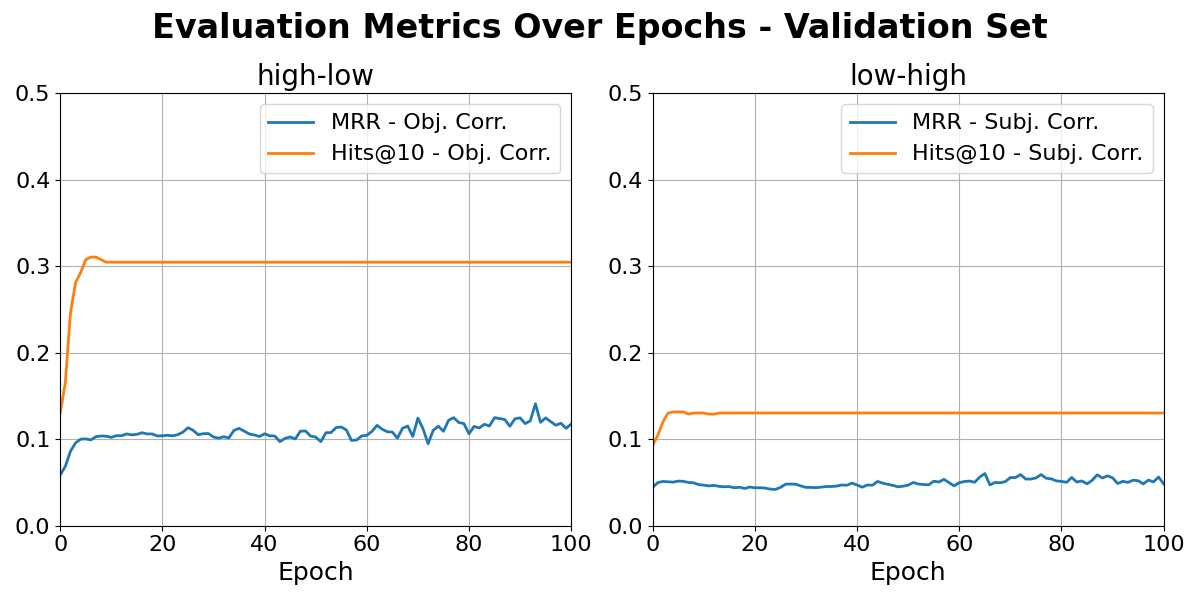}
        \caption{RotatE}
        \label{fig:no_degeneration_rotate}
    \end{subfigure}
    % --- ComplEx ---
    \begin{subfigure}{0.48\linewidth}
        \centering
        \includegraphics[width=\linewidth]{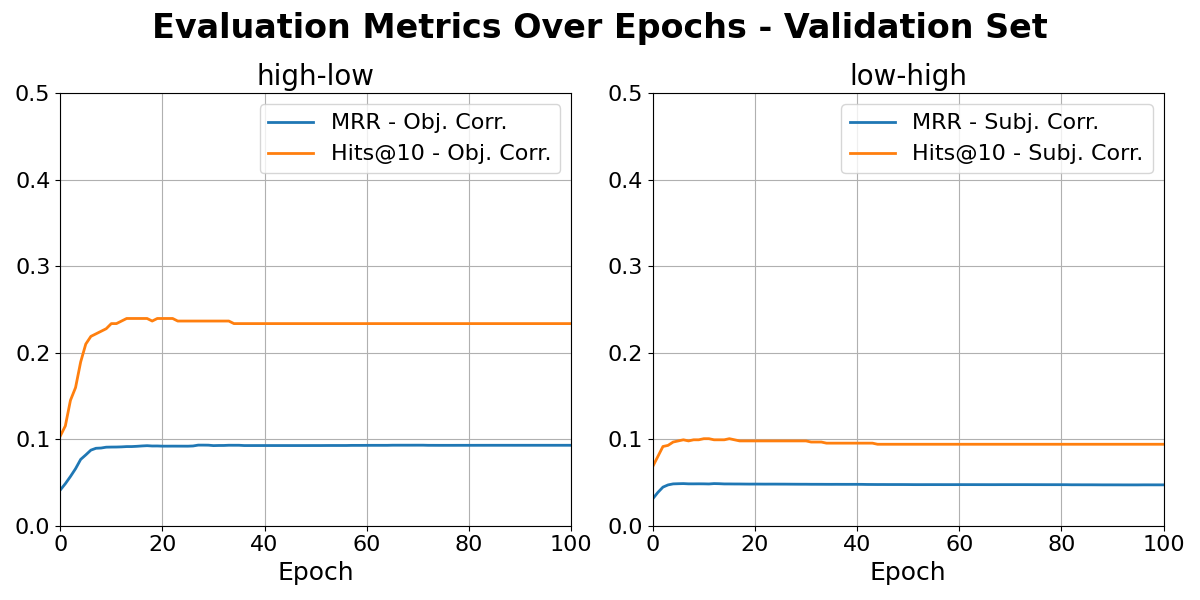}
        \caption{ComplEx}
        \label{fig:no_degeneration_rotate}
    \end{subfigure}

    \caption{Validation metrics for FB15k-237 extending the latent search to 100 epochs. Metrics plateau after reaching convergence without any collapse, showing how the optimization process guarantees no degenerate collapse of the embeddings.}
    \label{fig:no_degeneration}
\end{figure*}

The loss function is purely maximization over a set of known positive facts ($\mathcal{C}_q$) and oracle-suggested positive facts ($\Omega_q$). This might lead to questioning the stability of the optimization process, with the risk of collapse or severe overfitting. We did not observe such behaviour, as the design of ImbalancE prevents it for two reasons:
\begin{itemize}
    \item The first term of the loss function is used to optimize only the embedding of the anchor node, while the embeddings of the target entities of the training context remain fixed, thus preventing a collapse of the embeddings.
    \item{Similarly, in both terms of the loss function, the embedding of the relation is kept constant, providing a similar constraint that prevents an excess of overfitting}.
\end{itemize}
To further validate such a claim, we ran ImbalancE on top of our models for 100 epochs for FB15k-237. As evident from Figure~\ref{fig:no_degeneration}, MRR and H@10 on the validation set plateau after the first few epochs and then remain pretty much constant, showing non-degenerate behavior. Moreover, we remark once more how ImbalancE is meant to be used at inference time: as such, we expect to run it for a limited number of iterations, as plots below show is recommendable.

\subsubsection{Training Context Size v Performance} 
\begin{figure*}
    \centering
    \includegraphics[width=0.8\linewidth]{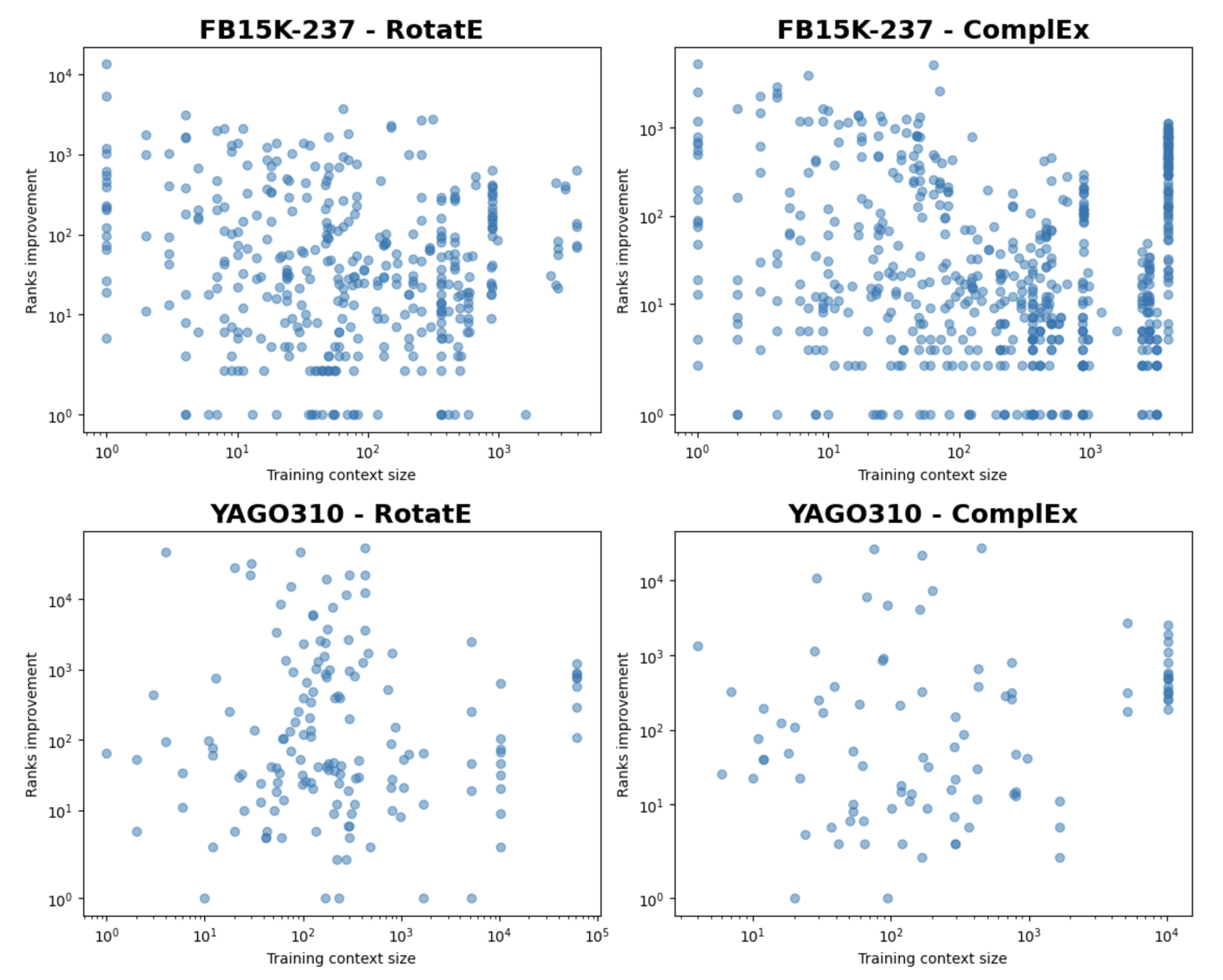}
    \caption{Correlation between the size of the training context and the rank improvement in FB15k-237 and Yago3-10.}
    \label{fig:correlation_context_size}
\end{figure*}

We try to understand whether there is any correlation between the size of the training context and the improvement that \textsc{ImbalancE} achieves on imbalanced triples. However, Figure~\ref{fig:correlation_context_size} shows how there is no such correlation, showing that the method can benefit imbalanced triples independently of the amount of information available in the training data, further supporting degree imbalance, rather than training-context size, as the salient factor here.

\newpage

\subsubsection{Execution with different seeds} Our method exhibits little to no variability across random seeds. The optimization starts from the fixed embeddings of the pretrained KGE model rather than from a random initialization, and the objective is evaluated on a fixed set of observed triples with no negative sampling, no stochastic mini-batching, and no dropout. Also the Adam optimizer contributes no randomness: its moment estimates are initialized deterministically to zero, so, given identical gradients, it produces identical updates. The only variability is the hardware-dependent floating-point non-determinism on GPU. This, however, is extremely limited, and in fact we have registered exactly the same performance on all datasets averaging across five different seeds.

\end{document}